%% file: main.tex
\documentclass[]{fairmeta}

\usepackage{wrapfig}
\usepackage{tabularx}
\usepackage{textcomp}
\usepackage{stfloats}
\usepackage{url}
\usepackage{verbatim}
\usepackage{titlesec}
\usepackage{adjustbox}
\usepackage{tikz}
\usepackage{comment}
\usepackage{amsmath,amsfonts,amssymb}
\usepackage{colortbl}
\usepackage{color}
\usepackage{hyperref}
\usepackage{graphicx}
\usepackage{subcaption} 
\usepackage{multirow}
\usepackage{booktabs}
\usepackage{xcolor}
\usepackage{pifont}
\usepackage{bbding}
\usepackage{fontawesome}
\usepackage{xspace}
\usepackage{float}
\usepackage{algorithm}
\usepackage{listings}
\usepackage{algcompatible}
\usepackage{algpseudocode}
\usepackage{soul}
\usepackage{array}
\usepackage{tcolorbox}
\usepackage{diagbox}
\usepackage{makecell}
\usepackage{rotating}
\usepackage{CJKutf8}
\usepackage{forest}
\usepackage{booktabs}
\usepackage{multirow}
\usepackage{caption}

\input{math_commands.tex}

\RequirePackage{xspace}
\makeatletter
\DeclareRobustCommand\onedot{\futurelet\@let@token\@onedot}
\def\@onedot{\ifx\@let@token.\else.\null\fi\xspace}

\makeatother

\definecolor{lightgrey}{RGB}{201,203,209}

\definecolor{token_blue}{RGB}{84, 120, 140}

\newlength\savewidth

\newcolumntype{x}[1]{>{\centering\arraybackslash}p{#1pt}}
\newcolumntype{y}[1]{>{\raggedright\arraybackslash}p{#1pt}}
\newcolumntype{z}[1]{>{\raggedleft\arraybackslash}p{#1pt}}

\renewcommand{\paragraph}[1]{\vspace{1mm}\noindent\textbf{#1}}

\renewcommand{\paragraph}[1]{\vspace{1.25mm}\noindent\textbf{#1}}

\definecolor{codeblue}{rgb}{0.25, 0.5, 0.5}
\definecolor{codekw}{rgb}{0.35, 0.35, 0.75}

\definecolor{ired}{RGB}{224,0,0}
\definecolor{ipurple}{RGB}{126,3,196}
\definecolor{igreen}{RGB}{3,196,144}
\definecolor{iorange}{RGB}{255,188,3}

\lstdefinestyle{Pytorch}{
    language = Python,
    backgroundcolor = \color{white},
    basicstyle = \fontsize{9pt}{8pt}\selectfont\ttfamily\bfseries,
    columns = fullflexible,
    aboveskip=1pt,
    belowskip=1pt,
    breaklines = true,
    captionpos = b,
    commentstyle = \color{codeblue},
    keywordstyle = \color{codekw},
}

\definecolor{green}{HTML}{009000}
\definecolor{red}{HTML}{ea4335}

\newcommand{\eqcontrib}{\clubsuit}
\newcommand{\correspond}{\spadesuit}

\newtcolorbox{promptblock}{
    colback=gray!5,
    colframe=gray!15,
    boxrule=0.5pt,
    arc=3pt,
    left=12pt,
    right=12pt,
    top=8pt,
    bottom=8pt,
    boxsep=8pt,
    breakable
}

\usepackage{paralist}
\setdefaultleftmargin{1.2em}{}{}{}{}{}

\title{\textsc{UniVA}: Universal Video Agent towards Open-Source Next-Generation Video Generalist}

\author[1,\eqcontrib]{Zhengyang Liang}
\author[2,\eqcontrib]{Daoan Zhang}
\author[3]{Huichi Zhou}
\author[4]{Rui Huang}
\author[4]{Bobo Li}
\author[5]{Yuechen Zhang}
\author[4]{\\Shengqiong Wu}
\author[6]{Xiaohan Wang}
\author[2]{Jiebo Luo}
\author[1]{Lizi Liao}
\author[4,\correspond]{Hao Fei}

\affiliation{$^1$Singapore Management University}
\affiliation{$^2$University of Rochester}
\affiliation{$^3$University College London}
\affiliation{\\$^4$National University of Singapore}
\affiliation{$^5$The Chinese University of Hong Kong}
\affiliation{$^6$Stanford University}

\contribution[\eqcontrib]{Core contributors, equal contribution (zyliang@smu.edu.sg; daoan.zhang@rochester.edu)}
\contribution[\correspond]{Project lead, correspondence (haofei7419@gmail.com)}

\input{abstract}

\date{\today}
\metadata[Website for Demos, Codes, Resources]{\url{http://univa.online/}}

\begin{document}
\thispagestyle{firstheader}
\maketitle
\pagestyle{fancy}
\fancyhf{}
\fancyfoot[C]{\thepage}


\section{Introduction}

\vspace{-1mm}

Real-world video applications often require composite, iterative, and \textit{agentic} workflows that go beyond any single AI capability~\citep{yu2023self,maazi2024video,song2024moviechat,fei2025path}.
For example, creating a dynamic visual story might begin with an image or text concept, expand into a generated video, then involve editing that video, segmenting key objects, and finally composing multiple elements into a polished scene. 
Traditionally, accomplishing such a pipeline requires stitching together disparate tools—each specialized for a narrow task—resulting in a brittle, labor-intensive, and non-interactive process that lacks automation and proactive assistance.
The lack of a unified system for reasoning across multiple video tasks and steps has become a critical bottleneck for next-generation video intelligence.

\begin{figure}[!th] 
\centering
\includegraphics[width=1\linewidth]{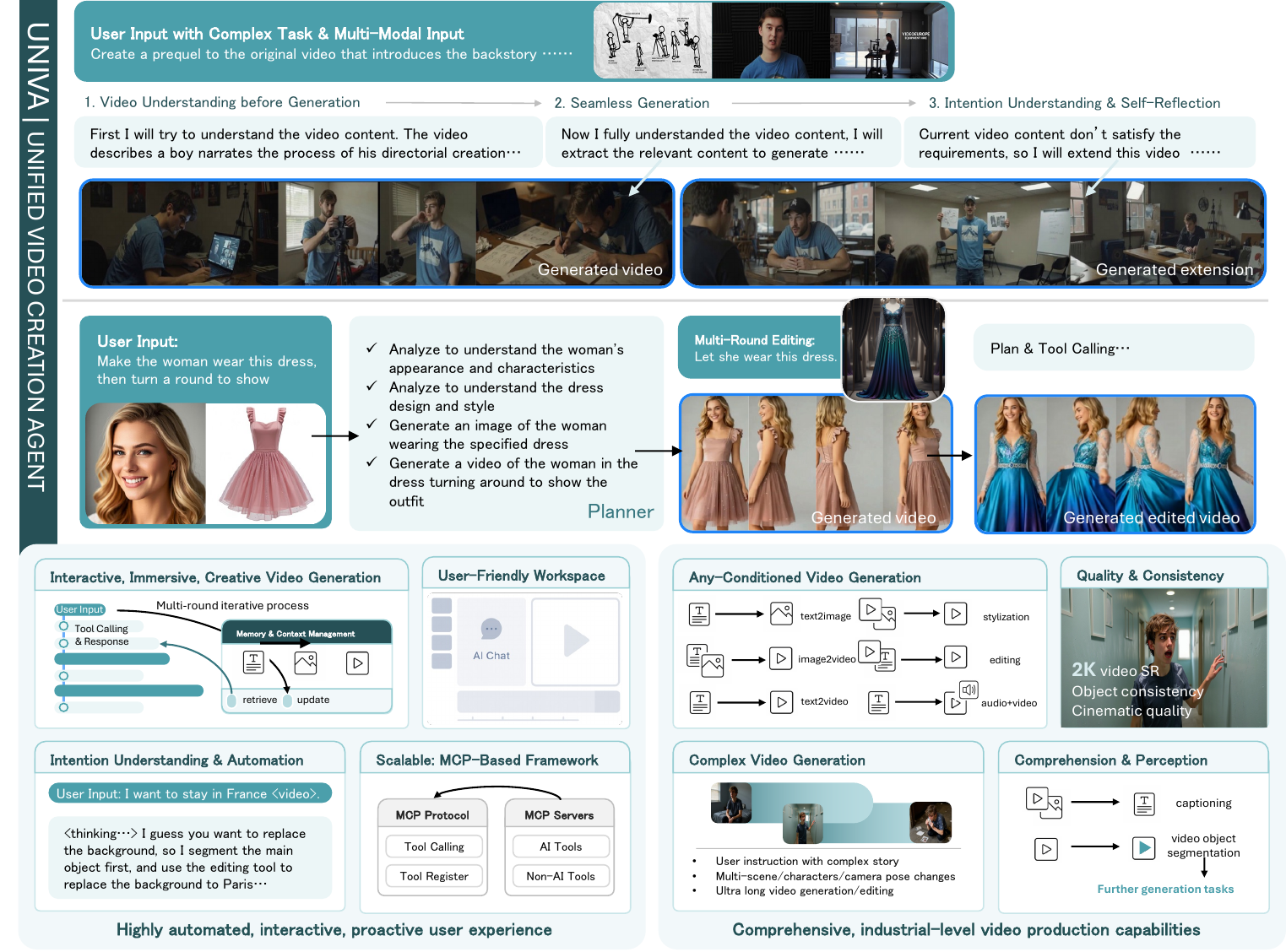}
\caption{UniVA (Universal Video Agent) delivers a highly automated, interactive, and proactive video creation experience, featuring multi-round dialogue co-creation, memory-based contextual reasoning, intent understanding, and tool-augmented planning for iterative user interaction.
It also serves as an omnipotent, unified, industrial-grade video production engine, integrating diverse generation, editing, and understanding modules within an MCP-based framework to ensure cinematic quality, consistency, and extensibility across any-conditioned video tasks.
}
\vspace{-3mm}
\label{fig:teaser}
\end{figure}

Existing approaches address parts of this challenge but fall short of a comprehensive solution. 
\textbf{Single-task video models} (\textit{e.g.}, dedicated networks for segmentation or video generation) deliver high performance on their specific tasks, yet they operate in isolation and fail to handle multi-step goals without manual coordination. 
More recently, \textbf{video-language foundation models} like VILA-U~\citep{wu2024vila} attempt to integrate understanding and generation into one model. 
These large models learn a broad spectrum of abilities \citep{fei2024vitron,xie2025show,tan2025omni}, but they remain monolithic and inflexible – they cannot easily incorporate new tools or modular functions, and leveraging them for complex workflows can be inefficient or impractical. 
Another emerging direction is to use \textbf{LLM-based agents with tool use}. 
For instance, HuggingGPT~\citep{shen2023hugginggpt} employs a language model as a controller to plan tasks and invoke appropriate models/tools in sequence. 
Similarly, VideoAgent~\citep{fan2024videoagent} leverages an LLM with a structured memory and a predefined set of video tools to answer questions on long videos. 
These agent-based systems illustrate the power of planning and tool integration \citep{Kugo25VideoMultiAgents,Wei25PreMind}. 
However, HuggingGPT is a generalist framework not specialized for detailed video operations, and VideoAgent focuses mainly on video understanding queries (\textit{e.g.}, Q\&A) with limited editing or generation capabilities. 
To date, no existing platform fully supports a unified, end-to-end \textit{agentic video workflow} that seamlessly integrates understanding, generation, and editing with proactive interaction and extensible modularity.

To bridge this gap, we propose \textbf{UniVA} (\textbf{U}niversal \textbf{V}ideo \textbf{A}gents), a unified multi-agent video AI platform that enables complex multi-step video creation and manipulation tasks. 
UniVA serves as both a \textit{creative video agent} for dynamic, user-interactive generation, and an \textit{industrial-grade video engine} for comprehensive, high-quality production.
Technically, UniVA can be depicted by two key characters:
\begin{itemize}
    \item \textbf{Highly automated, interactive, and proactive video creation experience:}
    UniVA is built on a \textit{Plan/Act dual agent architecture}: a planner agent first interprets the user's request and decomposes it into a sequence of actionable steps, and an executor agent (or a team of specialized agents) then carries out each step by invoking the appropriate video tool modules. 
    This separation of planning and acting (in line with recent agent design patterns) allows the system to look ahead and reason about long-horizon goals, while flexibly adapting the plan if intermediate results require changes. 
    On one hand, with strong planning capabilities, UniVA can autonomously accomplish an entire video production pipeline from a single user query.  
    On the other hand, agents communicate and share information through a \textit{multi-level memory mechanism}: a global memory stores persistent knowledge and facts (\textit{e.g.} general video facts or precomputed embeddings), a task-specific memory retains context and intermediate results for the current workflow, and a user memory keeps track of user preferences or historical interactions. 
    This hierarchical memory design ensures that context is maintained throughout the workflow, enabling continuity and avoiding forgetting important details mid-task. 
    In this way, UniVA also supports iterative, multi-round interactions, enabling deeply immersive and dynamic creative experiences.
    Moreover, UniVA demonstrates strong \textit{implicit intent understanding} and context retention, enabling agents to proactively refine or suggest creative improvements during generation.

    \item \textbf{Omnipotent, unified, industrial-level video production capabilities:}  
    Built upon the Model Context Protocol (MCP)~\citep{mcp}, UniVA can seamlessly integrate state-of-the-art video functional modules—either open-source or API-based—in a plug-and-play fashion, where each tool module is implemented as a modular server and the two agents act as the client.  
    The tool hub spans three major categories: video tools (e.g., generation, understanding, editing), non-video tools (e.g., audio or image operation), and non-AI tools (e.g., video cutting).  
    This broad coverage encompasses nearly all functionalities required in the video production process.  
    For example, UniVA supports video generation/transformation from arbitrary conditions, e.g., text, image or video.  
    By combining with cutting-edge external video generation models, UniVA enables cinematic-quality production of long, complex, and narrative-rich videos.  
    Under the MCP framework, the system can also be effortlessly extended to incorporate new tools and capabilities, forming an open ecosystem that continuously evolves toward a truly unified, omni-capable video generalist.
    
\end{itemize}

To evaluate such systems, we release \textit{UniVA-Bench}, a suite of multi-step video tasks spanning understanding, segmentation, editing, and generation. 
Tasks are specified as goal cards with gold artifacts (\textit{e.g.}, evidence spans, masks, EDLs) and scored with both task metrics and agentic metrics (plan quality, tool-routing efficiency, memory use, trace completeness). 
UniVA-Bench is designed to test compositionality, tool swaps, and long-form reasoning—not just per-task accuracy. Code, benchmark, and evaluators are all open-sourced.

In summary, our contributions are threefold:

\textbf{(1)} We present \textbf{UniVA}, the first \textit{open-source, unified, and omni-capable video generalist framework}.  
UniVA unifies video understanding, editing, and generation into a single agentic workflow, built upon a \textit{Plan–Act dual-agent architecture} with hierarchical multi-level memory.  
This design enables a \textit{highly automated, interactive, and proactive video creation experience} that dynamically reasons over long-horizon goals and user intent.  

\textbf{(2)} We develop a \textit{modular, MCP-based video production engine} that connects diverse open-source or API-based tools under a plug-and-play ecosystem.  
This framework supports any-conditioned generation, cross-modal editing, and industrial-level quality control, ensuring \textit{cinematic consistency, scalability, and extensibility} across all stages of video production.  

\textbf{(3)} We release \textbf{UniVA-Bench}, the first benchmark for evaluating \textit{agentic video intelligence} across multi-step compositions.  
It measures not only task performance but also planning quality, memory utilization, and tool-routing efficiency, offering a principled foundation for developing truly general-purpose video agents.

\section{Related Work}

\paragraph{Video Processing and Intelligence.} 
Videos serve as realistic simulations of the physical world, and research on video intelligence has spanned tasks such as action recognition, event detection, captioning, retrieval, and video question answering~\citep{tran2015c3d, xu2015discriminative,yang2023vid2seq,wu2023cap4video,le2020hierarchical}. 
With the advent of large language models (LLMs), video understanding has advanced toward instruction-following and long-context reasoning~\citep{maaz2024video, lin2024video,fei2024video}, while fine-grained tasks such as segmentation, grounding, and object parsing offer pixel-level insights~\citep{cheng2023tracking,lei2020tvqa,jin2022embracing}. 
In parallel, video generation has progressed from autoregressive models, such as VideoGPT~\citep{yan2021videogpt}, to diffusion-based methods, including Imagen Video~\citep{ho2022imagen}, Make-A-Video~\citep{singer2022make}, and Runway Gen-2~\citep{runway2023gen2}, which have achieved improved fidelity and temporal coherence. 
Recent work also explores controllable synthesis via conditional inputs~\citep{ni2023conditional, ma2024follow,liu2025javisdit,wu2025any2caption} and semantic-level video editing through diffusion- and instruction-driven pipelines~\citep{wu2023tune,khachatryan2023text2video,liu2024video}. Despite remarkable progress, most systems remain fragmented and task-specific, which limits their interoperability and scalability.

\vspace{-2mm}

\paragraph{Toward Unified Video Modeling.}  
To address fragmentation, unified video foundation models aim to integrate diverse tasks into a single framework. 
Joint models such as Show-o2~\citep{xie2025show} and Omni-video~\citep{tan2025omni} combine understanding and generation within large-scale multimodal training. Extensions of Video-LLMs~\citep{lin2024video,jin2024chat} integrate segmentation modules (e.g., SAM2~\citep{ravisam}) to support object-level grounding and reasoning~\citep{xiao2024can,yuan2025sa2va}, while modular architectures like \textsc{Vitron}~\citep{fei2024vitron} adopt flexible encoders and decoders for comprehension, segmentation, and generation. 
Although these efforts represent significant progress, most existing systems rely on static pipelines without effective scheduling or coordination mechanisms, making them difficult to extend. This highlights the need for frameworks that facilitate dynamic orchestration of heterogeneous modules.

\vspace{-2mm}

\paragraph{Agents for Video Intelligence.}  
Agent-based paradigms have emerged as a promising solution for flexible video intelligence, leveraging planning, interaction, and memory mechanisms~\citep{Chen24ShareGPTVideo,Yin23SML,Wu24NExTGPT}. \textsc{VideoAgent}~\citep{Fan24VideoAgent} enhances generative quality with memory augmentation, while other works explore agent planning for long-context reasoning~\citep{Wang24VideoAgent} and self-improving generation~\citep{Soni24VideoAgent}. Applications extend to video reasoning~\citep{Liu25VideoMind,Shi25EVR}, editing~\citep{Wang24LAVE}, stylization~\citep{Yue25VStylist}, and story generation~\citep{Hu24StoryAgent}. Multi-agent collaborations such as VideoMultiAgents~\citep{Kugo25VideoMultiAgents} and \textsc{PreMind}~\citep{Wei25PreMind} further enhance performance, though communication and coordination remain open challenges. Protocols like \textsc{MCP}~\citep{mcp} and modular plug-and-play designs offer promising directions. Departing from isolated paradigms, our UniVA framework leverages multi-agent interaction, memory augmentation, and context engineering~\citep{Mei25SCE} to unify understanding, reasoning, editing, and generation, advancing toward truly universal video agents.

\begin{figure}[!t]
  \centering
   \includegraphics[width=1\linewidth]{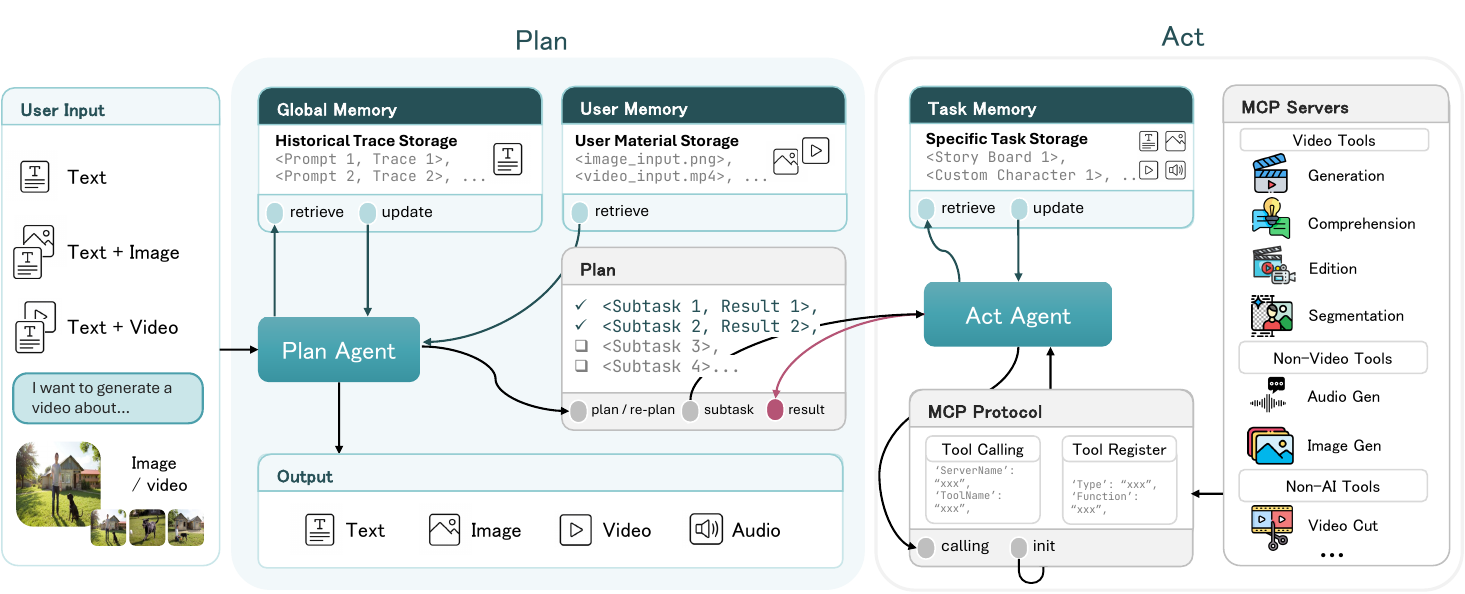}
   \vspace{-6pt}
   \caption{
   Overall architecture of the proposed UniVA system, built on a Plan–Act paradigm.
   The Plan Agent decomposes user input (text, image, or video) into subtasks by leveraging global memory (historical traces) and user memory (stored materials).
   The Act Agent retrieves task-specific memory, executes subtasks via the MCP protocol, and coordinates with external MCP servers (video, AI, and non-AI tools). The system generates multimodal outputs, including text, image, video, and audio.
   }
   \vspace{-8pt}
   \label{fig:framework}
\end{figure}

\section{UniVA}

\vspace{-2mm}
\subsection{Problem Formulation}

\vspace{-2mm}
We formulate the challenge of synergistic video intelligence as a sequential decision-making problem. The UniVA agent operates within an environment defined by a user's high-level goal $G$ and a set of available tools $\mathcal{T}$. The agent's objective is to generate a sequence of actions $A = (a_1, a_2, \dots, a_N)$ that transform an initial state $s_0$ (which may include user-provided videos or images) into a final state $s_N$ that satisfies the goal $G$.

At each step $t$, the agent, guided by a policy $\pi$ (instantiated by our Planner), observes the current state $s_t$ and the interaction history $H_t = (a_1, s_1, \dots, a_{t-1}, s_{t-1})$, and selects an action $a_t \in \mathcal{T}$. The execution of this action by the Actor transitions the environment to a new state $s_{t+1} = \text{Execute}(s_t, a_t)$. The Memory system serves as the persistent representation of the history $H_t$ and intermediate artifacts within the state $s_t$.

The core challenge, therefore, is to design an agent with a policy $\pi$ that can effectively manage the state transitions and leverage the history to produce a high-quality final state $s_N$, demonstrating both breadth (by utilizing a large and diverse $\mathcal{T}$) and depth (by creating complex, synergistic action sequences $A$). This entire process can be summarized as finding the optimal action sequence $A^*$ that maximizes a quality function $\mathcal{Q}$ of the final state with respect to the goal:
\begin{equation}
\label{eq:optimization}
A^* = \underset{A=(a_1, \dots, a_N)}{\operatorname{argmax}} \mathcal{Q}(s_N, G)
\quad \text{where} \quad
s_{t+1} = \text{Execute}(s_t, a_t)
\end{equation}

\subsection{The UniVA Control Core}

To achieve the aforementioned sequential decision-making framework, we designed the control core of UniVA. This core consists of two key components: a decision engine responsible for formulating the policy $\pi$, and a memory system tasked with managing the state $s_t$ and history $H_t$.

\subsubsection{Plan--Act Dual Agent Architecture}

\vspace{-2mm}

The core of UniVA is a Plan--Act dual-agent architecture. 
Our strategy $\pi$ is implemented through a dual-agent Plan-Act framework. The Planner, as a high-level policy network, maps the goal  
$G$ and the current state $s_t$ to an abstract sequence of plan. The Actor, as a low-level execution policy, converts each step of the plan into concrete actions $a_t$.

For example, given ``make a cartoon video of my dog,'' the Planner may decompose it into: (1) retrieve images of the dog, (2) generate a cartoon-style video, (3) edit the background, and (4) compose audio. 
The \textbf{Actor} receives each sub-goal from the Planner, selects the appropriate tool through the MCP interface\footnote{\url{https://modelcontextprotocol.io/}} \citep{mcp}, fills in required arguments (e.g., video clip, mask, prompt), and executes the call. 
Once a tool finishes, the Actor collects the output and sends it back to the Planner.  
This separation keeps the Planner lightweight and strategic, while the Actor focuses on reliable and efficient tool use.

\subsubsection{Memory Mechanism}

\vspace{-2mm}

\begin{figure}[t!]
  \centering
   \includegraphics[width=1\linewidth]{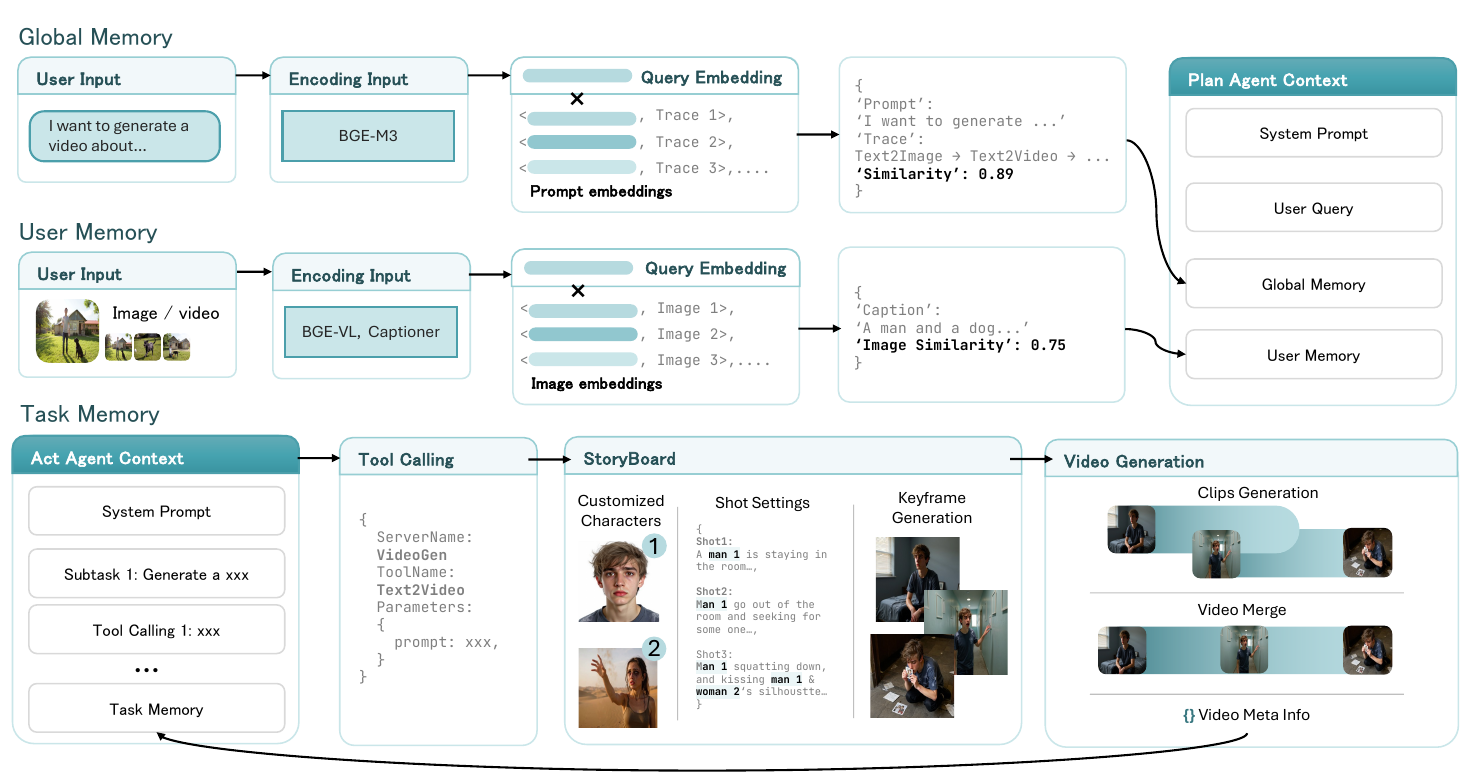}
   \vspace{-6pt}
   \caption{
   Memory-augmented framework for video generation.
   Global and user memories provide context to the plan agent, while task memory coordinates tool calling, storyboard creation, and video generation.
   }
   \vspace{-7pt}
   \label{fig:memory}
\end{figure}

\begin{figure}[t!]
  \centering
   \includegraphics[width=1\linewidth]{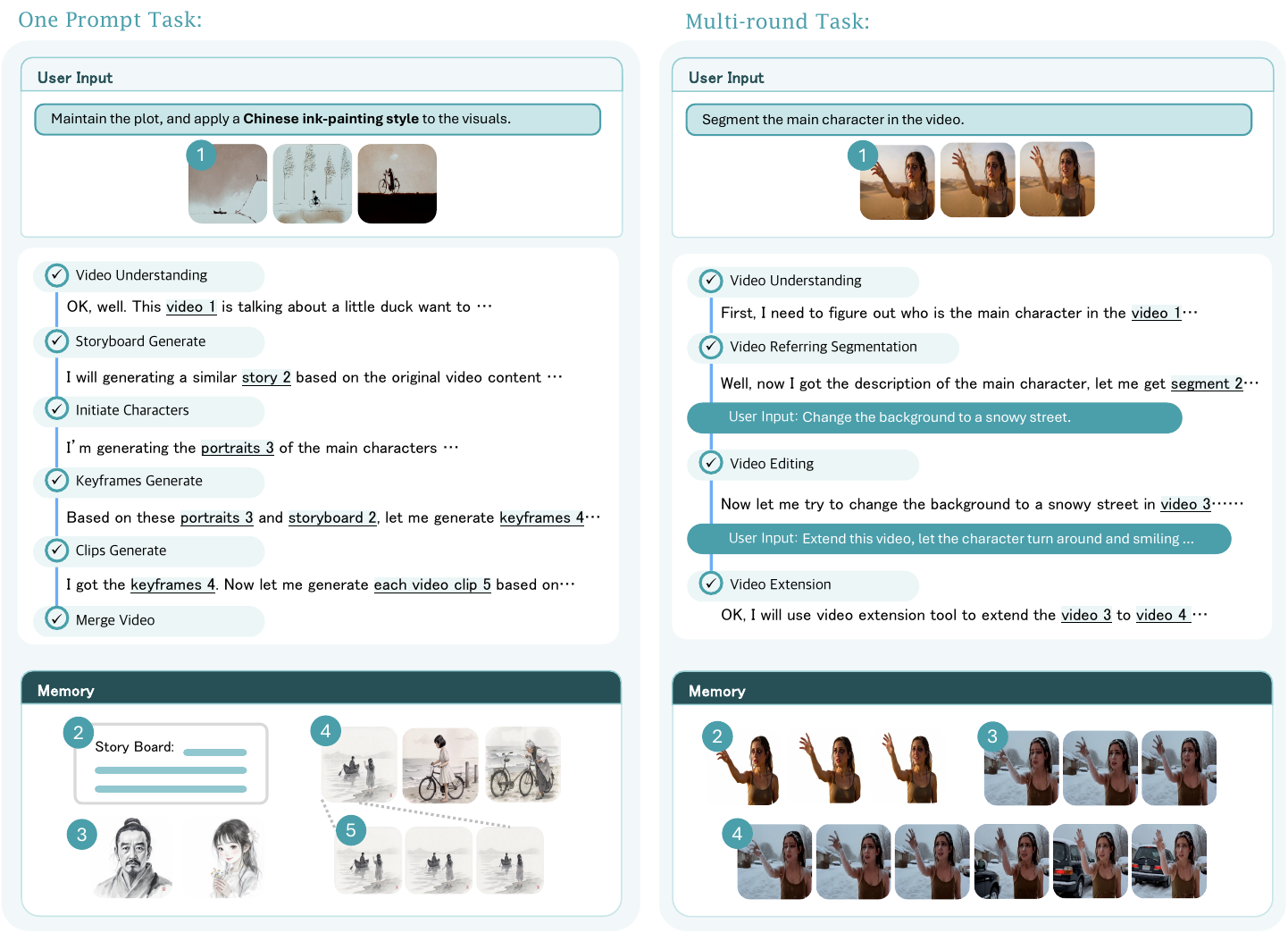}
   \vspace{-6pt}
   \caption{
   Iterative tool calling for video generation in UniVA. Left: one-prompt task applies a global ink-painting style. Right: multi-round task incrementally edits via segmentation, background change, and extension, demonstrating representative functions.
   }
   \vspace{-7pt}
   \label{fig:process}
\end{figure}

A key challenge in agentic video systems is to maintain context across long and multi-step workflows.  
As presented in Figure \ref{fig:memory}, UniVA addresses this with a \emph{three-level memory mechanism} that complements the Planner--Actor loop:
\textbf{Global Memory. } 
      Stores persistent knowledge and reusable resources, such as precomputed embeddings, generic video facts, or tool usage statistics.  
      This memory provides background context and supports cross-task generalization.
\textbf{Task Memory.}  
      Maintains intermediate artifacts, tool outputs, and execution traces for the current workflow.  
      It ensures continuity across multiple steps, allowing later sub-goals to reuse results (e.g., segmentation masks or captions) without recomputation.  
      Task memory also enables traceability, making the entire workflow transparent and reproducible.
\textbf{User Memory.}  
      Tracks user-specific preferences and historical interactions, such as favored styles, recurring edit patterns, or personalized constraints.  
      This enables adaptive behaviors—for example, automatically applying a user’s preferred resolution or editing style in future tasks.

Through this design, Global Memory and User Memory together form the persistent storage of long-term history $H_t$, providing rich context for the strategy $\pi$. Task Memory maintains the dynamic state of the current task $s_t$ and its intermediate products.


\subsection{Tools Interaction}

\vspace{-2mm}

The capability of an agent ultimately depends on its available action space, i.e., the toolset $T$. To achieve maximum breadth, UniVA's action space is designed to be open and extensible.

We achieve unified management of the action space $T$ through MCP protocol. The MCP server module acts as a unified gateway between the Actor and a collection of distinct tool servers. The server maintains a registry of available functions, validates and executes calls through a standardized API, and records outputs for traceability. This design means that adding or replacing a capability only requires registering it on the server, while the Planner and Actor remain unchanged, making the system modular and extensible. The detailed tools list at \ref{sec:functions walkthrough}.

\subsection{Framework Walkthrough}

Building upon the above components, we now present the overall architecture of our framework. This sequential decision-making process is illustrated in Figure~\ref{fig:framework}. 
In any task, the Planner (policy $\pi$) observes the current state $s_t$ and the goal $G$ to formulate a plan, which the Actor executes as a sequence of actions $A=(a_1, \dots, a_N)$ using various tools. 
The Memory will record the outputs of each action, continuously updating the history $H_t$ and the state $s_t$ for subsequent steps.

The two cases demonstrate the versatility of this framework. 
The left case exemplifies the agent's depth, autonomously decomposing a single complex goal into a coherent, multi-step generation workflow. 
The right case showcases the interplay of breadth, where a diverse and extensible toolset $\mathcal{T}$, made accessible via our MCP layer, is synergistically chained in an interactive session to achieve a precise, context-dependent editing outcome.

\subsection{System Implementation}

\begin{figure}[t!]
  \centering
   \includegraphics[width=1\linewidth]{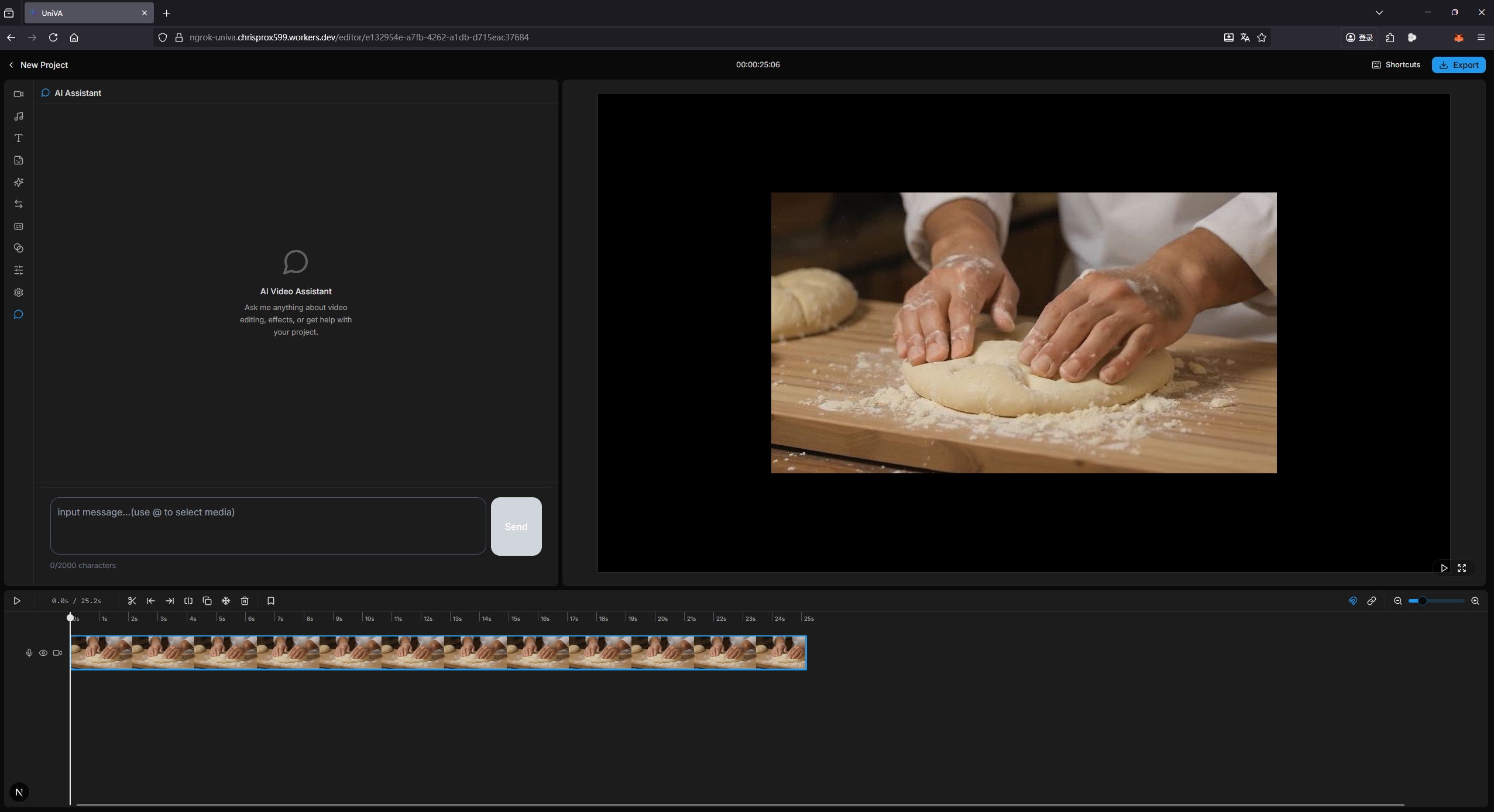}
   \vspace{-6pt}
   \caption{
   The interface combines a traditional non-linear timeline and preview canvas with a conversational assistant (left), which provides a user-friendly entry point to the UniVA agent. This design supports both one-stop, prompt-based generation and multi-turn, interactive editing workflows.
   }
   \vspace{-6pt}
   \label{fig:web}
\end{figure}

To demonstrate the practical application of our framework, we have instantiated the UniVA agent within an interactive, web-based video editing application\footnote{Our frontend system is built on \href{https://github.com/OpenCut-app/OpenCut}{OpenCut}, and we appreciate the open-source contributions made by the OpenCut team.}, as shown in Figure \ref{fig:web}.

The UniVA agent, operating in the background, parses these requests, formulates a plan, and executes the necessary tool calls. The results are then directly reflected on the timeline and preview canvas. This tight integration creates a fluid, iterative loop, allowing users to effortlessly switch between high-level AI-driven creation and traditional, hands-on editing, all within a single, unified platform.

\begin{figure}[t!]
  \centering
   \includegraphics[width=1\linewidth]{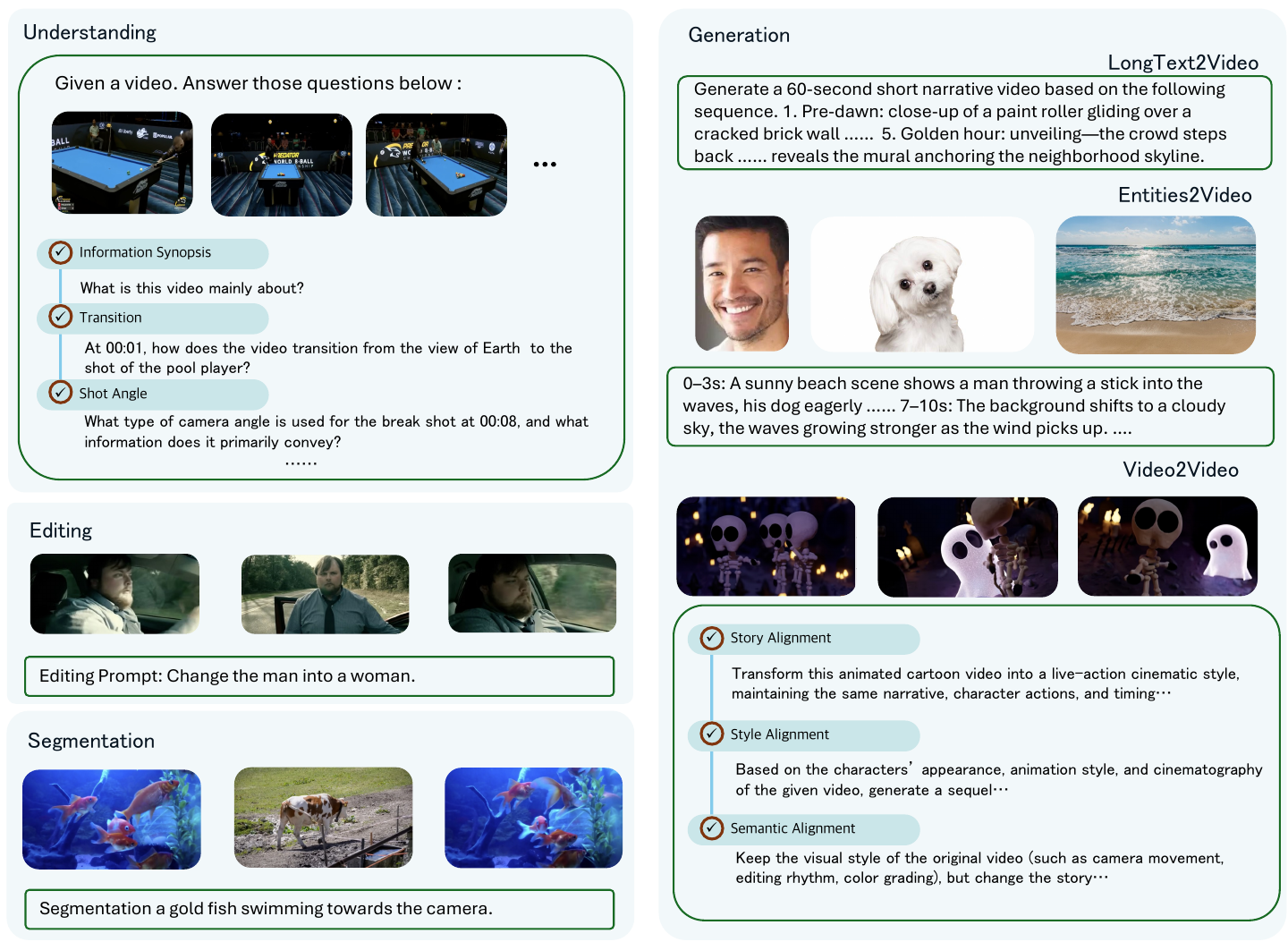}
   \vspace{-6pt}
   \caption{
   Representative examples from the four main task categories in UniVA-Bench: Understanding, Generation, Editing, and Segmentation.
   }
   \vspace{-6pt}
   \label{fig:benchmark}
\end{figure}

\section{UniVA-Bench}

\subsection{Benchmark Definition}

Video intelligence in practice is an iterative, multi-stage creation process where users interleave understanding, generation, editing, segmentation, and audio/asset composition within a single workflow. However, most existing benchmarks largely isolate single tasks and single models, which underestimates the difficulty of long-horizon, multi-step video production and the need for explicit planning, memory, and tool orchestration. Therefore, we introduce a unified agent-oriented benchmark that shifts the focus from isolated single-model tasks to end-to-end, tool-augmented video intelligence, aligning evaluation with real user workflows and the requirements of practical video agents.

To holistically assess both the range and the intelligence of an agent, the benchmark is organized into two complementary tracks:
i) Functional Modules: task performance across Understanding, Generation (LongText2Video, Image/Entities2Video, Video2Video), Editing (long video edits with cross-shot consistency), and Segmentation (long video segmentation with multi-entity occlusion).
ii) Agentic Probing: plan quality, dependency satisfaction, and re-planning robustness using structured plan-level metrics; analysis of memory usage (trace, user, task/storyboard) and its downstream impact.

\subsection{Evaluation Tasks}

\textbf{Understanding (Long-Video QA).}
This task is designed to target both aesthetics- and semantics-oriented questions for long videos, encompassing shot transitions, visual style, and narrative comprehension in addition to standard entity and action semantics. Unlike prior settings, where each QA pair is tied to a single short clip, our task demands answering multiple interdependent questions grounded in a single long-form video.

\textbf{Generation.}
Agents are evaluated on diverse real-world video generation tasks, categorized into three subtypes: \textit{1) LongText2Video}, handling long or noisy prompts that necessitate storyboard-first planning; \textit{2) Image/Entities2Video}, using 1--3 reference images to enforce identity preservation and cross-scene coherence; \textit{3) Video2Video}, conditioning on a source video while ensuring referential stability for persons and objects.

\textbf{Editing (Long Video).}
This task is defined to involve multi-step edits such as cross-shot replacement, attribute modification, and style transfer, while maintaining narrative integrity and referential consistency. Effective completion requires reasoning in combination with tool invocation (e.g., ref-seg $\rightarrow$ inpaint/compose $\rightarrow$ merge). 

\textbf{Segmentation (Long Video).}
Designed for long clips with multiple entities and frequent occlusions, this task evaluates temporal consistency and robustness in detecting and segmenting shot boundaries.  

\textbf{Agentic probing sets.}
We include
(1) a 50-instance storyboard$\rightarrow$user-intent planning set to compare Single-Agent vs. Plan-Act, and
(2) a set of standard pipeline tasks with expert references to assess wPED, DepCov, and ReplanQ under injected failures.
Memory analyses consider \emph{trace memory} (historical trajectories), \emph{user memory} (preferences), and \emph{task memory} (e.g., storyboards).

More data curation details in Appendix \ref{sec:curation}.

\subsection{Evaluation Protocol}
To evaluate agent performance on UniVA-Bench, we employ a comprehensive suite of metrics targeting three key areas: (1) Task-specific Quality, using established metrics like CLIP Score for command following and DINO Score for subject consistency; (2) Overall User Preference, captured via pairwise judgments from a powerful MLLM-as-a-Judge; and (3) Agentic Planning Capabilities, assessed using our novel, specialized metrics (wPED, DepCov, and ReplanQ) that measure plan quality, logical correctness, and recovery robustness. The detailed definitions and calculation methods for all metrics are provided in Appendix \ref{sec:metrics}.

For Generation/Editing, we report CLIP, DINO, and MLLM preference.
for Segmentation, J/F/J\&F;
for Understanding, normalized QA accuracy.
For agentic probing, we report wPED/DepCov/ReplanQ with and without memory (trace/user/task) and compare Single-Agent vs. Plan-Act frameworks.

\section{Experiments}

To comprehensively evaluate our system's capabilities in realistic, end-to-end workflows, we conduct all experiments on UniVA-Bench, a novel agent-oriented benchmark we introduce in this work.
Our experiments are designed to test two central hypotheses: i) a unified agentic architecture, where functional modules like understanding and generation are deeply integrated, provides a significant performance advantage over isolated, end-to-end models; and ii) the combination of a dual-agent Plan-Act framework and a multi-component memory system is essential for achieving the robust planning and persistent context required for complex video tasks. The complete experiment settings are in the Appendix \ref{sec:detailed_settings}.

\subsection{Performance of Functional Modules}

\subsubsection{Generation}

In the generation scenarios, we benchmark UniVA against three representative end-to-end models: LTX-Video ~\citep{HaCohen2024LTXVideo}, Wan ~\citep{wan2025wanopenadvancedlargescale}, and Seedance ~\citep{gao2025seedance10exploringboundaries}. Evaluating the results using CLIP Score (prompt following), DINO Score (subject consistency), and preference ratings from an MLLM-as-a-Judge, following the UniVA-Bench specification. Detailed baseline setups are in the Appendix.

\begin{table}[h!]
\centering
\caption{Comparison across LongText2Video, Entities2Video and Video2Video.}
\label{tab:generation}
\resizebox{\textwidth}{!}{%
\begin{tabular}{lccccccccc}
\toprule
\multirow{2}{*}{\textbf{Method}} & \multicolumn{3}{c}{\textbf{LongText2Video}} & \multicolumn{3}{c}{\textbf{Entities2Video}} & \multicolumn{3}{c}{\textbf{Video2Video}} \\
\cmidrule(lr){2-4} \cmidrule(lr){5-7} \cmidrule(lr){8-10}
 & CLIP Score & DINO Score & MLLM Judge & CLIP Score & DINO Score & MLLM Judge & CLIP Score & DINO Score & MLLM Judge \\
\midrule
LTX-Video & 0.2161 & \textbf{0.9392} & 1.125 & 0.2210 & 0.8452 & 1.281 & 0.2263 & \textbf{0.9943} & 2.123 \\
Wan       & 0.2028 & 0.6779 & 3.183 & \textbf{0.3106} & 0.7043 & 1.650 & 0.2632 & 0.9188 & 2.034 \\
Seedance  & 0.2157 & 0.8836 & 2.650 & 0.3039 & \textbf{0.8800} & \textbf{2.700} & \textbf{0.2684} & 0.9518 & 2.621 \\
\textbf{UniVA} & \textbf{0.2814} & 0.9026 & \textbf{3.333} & 0.2868 & 0.8796 & 1.789 & 0.2620 & 0.8939 & \textbf{4.068} \\
\bottomrule
\end{tabular}%
}
\end{table}

\textbf{LongText2Video.} In the LongText2Video scenario, UniVA's superior performance - achieving the highest CLIP score 0.2814 and the MLLM Judge score 3.333 - is directly attributable to its agentic framework. Unlike end-to-end models, UniVA's Planner first parses the noisy, long-term text to distill the core user intent into an optimal prompt. Overcoming a common shortage of traditional end-to-end models.

\textbf{Entities2Video.} In this task, which tests the agent's ability to maintain subject identity from reference images, the results are more nuanced. While specialized models like Seedance show strong performance in subject consistency (DINO Score), UniVA remains competitive. This highlights a current trade-off where our agent prioritizes overall instruction complexity and narrative coherence, a direction for future optimization.

\textbf{Video2Video.} In the Video2Video task, although UniVA does not lead in automated metrics such as the CLIP Score or DINO Score, it achieves a commanding MLLM Judge score of 4.068. This apparent discrepancy shows that UniVA's planner excels at interpreting and executing complex instructions (e.g., `modify the storyline while preserving the style'). This often requires understanding of the original video then provide concise prompt to generate new video, which will naturally lowers strict frame-level similarity (DINO score), but results in a final video that better fulfills the user's holistic intent.

\subsubsection{Understanding}
For the understanding task, we compare UniVA against several leading Large Multimodal Models, including GPT-4o~\citep{openai2024gpt4ocard}, Gemini 2.5 Pro~\citep{gemini2023modelcard}, InternVL3-38B~\citep{zhu2025internvl3exploringadvancedtraining}, and Qwen2.5-VL-72B~\citep{bai2025qwen25vltechnicalreport}. Performance is measured by the normalized QA accuracy score as defined in the UniVA-Bench protocol.

\begin{table*}[h!]
\centering
\hspace{-0.05\textwidth}
\begin{subtable}[t]{0.3\textwidth}
    \centering
    \small
    \caption{LongVideo Understanding}
    \label{tab:longvideo_understanding}
    \begin{tabular}{lc}
        \toprule
        \textbf{Method} & \textbf{Acc} \\
        \midrule
        GPT-4o         & 0.52 \\
        Gemini 2.5 Pro & 0.65 \\
        InternVL3-38B  & 0.75 \\
        Qwen2.5-VL-72B & 0.74 \\
        \textbf{UniVA} & \textbf{0.76} \\
        \bottomrule
    \end{tabular}
\end{subtable}
\hspace{-0.01\textwidth}
\begin{subtable}[t]{0.3\textwidth}
    \centering
    \small
    \caption{Long Video Editing}
    \label{tab:editing}
    \begin{tabular}{lccc}
        \toprule
        \multirow{2}{*}{\textbf{Method}} & \multicolumn{3}{c}{\textbf{Editing}} \\
        \cmidrule(lr){2-4}
        & CLIP & DINO & MLLM \\
        \midrule
        Vace           & 0.2258 & 0.6808 & 3.484 \\
        \textbf{UniVA} & \textbf{0.2280} & \textbf{0.7488} & \textbf{3.635} \\
        \bottomrule
    \end{tabular}
\end{subtable}
\hspace{0.05\textwidth}
\begin{subtable}[t]{0.3\textwidth}
    \centering
    \small
    \caption{Long Video Segmentation}
    \label{tab:segmentation}
    \begin{tabular}{lccc}
        \toprule
        \multirow{2}{*}{\textbf{Method}} & \multicolumn{3}{c}{\textbf{Segmentation}} \\
        \cmidrule(lr){2-4}
        & J & F & J\&F \\
        \midrule
        SA2VA          & 0.2076 & 0.0972 & 0.1524 \\
        \textbf{UniVA} & \textbf{0.3254} & \textbf{0.1680} & \textbf{0.2467} \\
        \bottomrule
    \end{tabular}
\end{subtable}
\caption{Comparison of UniVA on three long video tasks: Understanding, Editing, and Segmentation.}
\label{tab:all_three}
\end{table*}

In Table \ref{tab:longvideo_understanding}, our UniVA agent achieves the highest accuracy of 0.76. Prove the agent's ability to decompose the video and the complex query into manageable sub-tasks leads to a more accurate and holistic understanding than what a single inference from a base model can provide.

\subsubsection{Editing}
For long video editing, we compare against Vace ~\citep{jiang2025vaceallinonevideocreation}, a strong baseline for video editing tasks. Metrics include CLIP Score, DINO Score, and MLLM preference.

In Table \ref{tab:editing}, it can be observed that, in traditional non-unified setup, an editing model would be disconnected from a deep, continuous understanding of the video. UniVA bridges this gap. The agent first leverages the integrated Understanding module via the Probing tool to establish a persistent semantic context, allowing the agent to ground editing objects on long-term, cross-shot video to apply its editing actions.

\subsubsection{Segmentation}
In the challenging long video segmentation task, we use SA2VA ~\citep{yuan2025sa2va} as our primary baseline. We report the J-mean, F-mean, and J\&F-mean scores.

In Table \ref{tab:segmentation}, UniVA outperforms the best scores on all metrics. Because UniVA can query the co-located Understanding module to resolve ambiguities that are impossible to solve at the pixel level. For instance, when an object is occluded, the agent can ask the Probing tool: ``Based on the narrative context, is the object reappearing at timestamp X the same `blue car' from timestamp Y?'' This ability to dynamically leverage a powerful understanding module to inform a perception task like segmentation is a unique benefit of our integrated design.

\noindent
\begin{tikzpicture}
\node[
  draw=blue!45, fill=blue!6,
  rounded corners=2mm,
  line width=0.6pt,
  inner sep=6pt,
  align=left,
  text width=\linewidth
]{
These 4 experiments demonstrate that a unified agentic architecture is critical for advancing video intelligence. The superior performance of UniVA is not merely due to the quality of its individual modules but stems from the tight coupling and dynamic interplay between them.
};
\end{tikzpicture}

\subsection{Agentic System Probing}

\subsubsection{Planning Capability}

In this section, we probe the core agentic capabilities of our system, moving beyond output quality to analyze the underlying planning and memory mechanisms. We first validate our choice of a Plan-Act framework and its Planner LLM component.

\begin{figure}[h!]
    \centering
    \begin{minipage}[t]{0.45\textwidth}
        \centering
        \includegraphics[width=\textwidth]{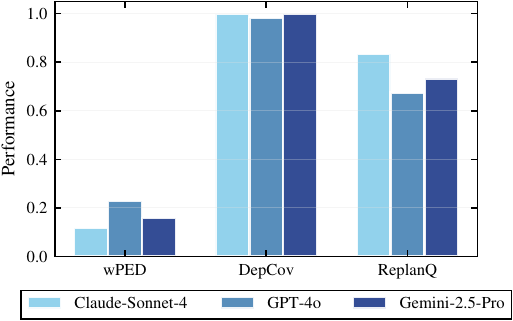}
        \caption{Performance of Planner LLMs.}
        \label{fig:left}
    \end{minipage}
    \hfill
    \begin{minipage}[t]{0.45\textwidth}
        \centering
        \includegraphics[width=\textwidth]{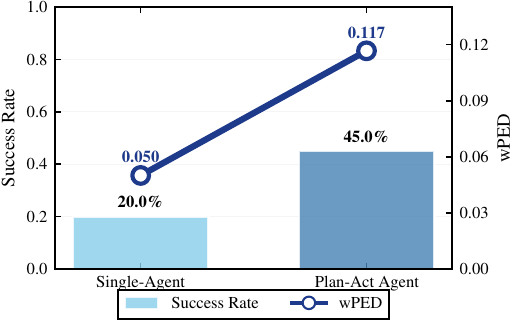}
        \caption{Framework comparison.}
        \label{fig:right}
    \end{minipage}
\end{figure}

To select the optimal Planner for our framework, we evaluated three leading LLMs on key agentic metrics (Figure \ref{fig:left}). Claude-Sonnet-4 demonstrated superior performance in DepCov and ReplanQ. Since correctly identifying task dependencies and robustly recovering from failures are paramount for a reliable agent, we selected Claude-Sonnet-4 as the Planner for all subsequent experiments.

In Figure \ref{fig:right}, Success Rate is the percentage of test cases where the agent produced a structurally valid plan (i.e., wPED > 0)—measuring the agent's ability to avoid catastrophic failures, such as generating an empty or malformed output. It more than doubles the Success Rate (45.0\% vs. 20.0\%), indicating a much lower rate of catastrophic failures. Furthermore, the quality of its successful plans is also over twice as high, reflected in a wPED score of 0.117 versus 0.050. This confirms that the explicit planning stage can not only output valid plans but also high-quality plans.

\subsubsection{Memory Capability}
We then analyze the distinct contributions of our three memory modules. To isolate their effects, we designed specific experimental probes: (i) Global Memory was tested by providing the agent with a set of trajectories from an expert planning dataset; (ii) User Memory was evaluated in the Entities2Video task, where the agent could retrieve user-provided reference images via a RAG mechanism; and (iii) Task Memory was assessed in the LongText2Video task by comparing the performance of generating with and without a storyboard.

\begin{figure}[h!]
    \centering
    \begin{minipage}[t]{0.3\textwidth}
        \centering
        \includegraphics[width=\textwidth]{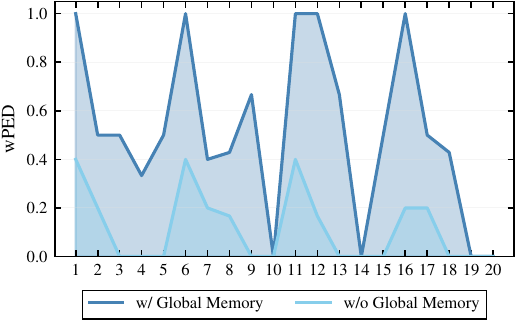}
        \caption{Effect of trace memory.}
        \label{fig:one}
    \end{minipage}
    \hfill
    \begin{minipage}[t]{0.3\textwidth}
        \centering
        \includegraphics[width=\textwidth]{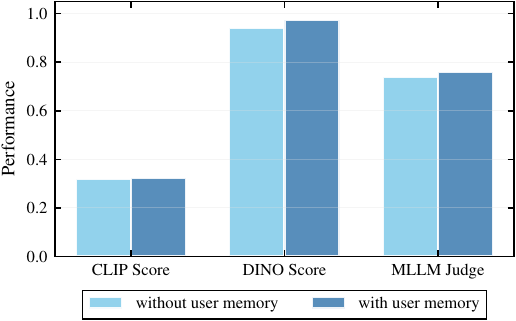}
        \caption{Effect of user memory.}
        \label{fig:two}
    \end{minipage}
    \hfill
    \begin{minipage}[t]{0.3\textwidth}
        \centering
        \includegraphics[width=\textwidth]{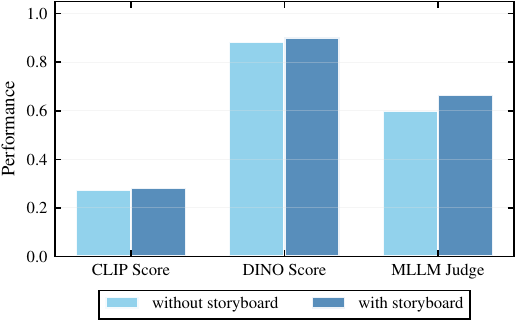}
        \caption{Effect of storyboard.}
        \label{fig:three}
    \end{minipage}
\end{figure}

In Figure \ref{fig:one}, across most cases, the agent with global memory (the dark blue line) achieves a higher wPED score than without global memory (the light blue line). This indicates that by drawing on past trajectories, the agent becomes better at aligning its generated plans with expert-preferred structures. And most strikingly, global memory prevents catastrophic planning failures. In numerous instances (e.g., turns 3-5, 8-10, 14, and 18-20), the agent without global memory completely fails to produce a viable plan, resulting in a wPED score of zero. However, agent with global memory not only succeeds but often produces high-quality plans. 
Figure \ref{fig:two} shows with the user memory, agent can better understand the indications of the user, such as when user refers to a cat, user memory can make agent has the ability to find the cat image from user's materials. Making the generated content more aligned with user intent.
Utilizing storyboards as task memory (Figure \ref{fig:three}) provided a substantial boost across all quality metrics. This demonstrates that maintaining an intermediate representation of the creative goal is essential for ensuring semantic coherence and cross-shot consistency in the final video, directly validating the storyboard's role in our agent's workflow.

\noindent
\begin{tikzpicture}
\node[
  draw=blue!45, fill=blue!6,
  rounded corners=2mm,
  line width=0.6pt,
  inner sep=6pt,
  align=left,
  text width=\linewidth
]{
In summary, our dual Plan-Act Agent framework improves the ability to process complex tasks. Also, three memory mechanisms help the agent to build a persistent context, to be more robust, better user intent understanding, and more consistent in generated videos.
};
\end{tikzpicture}  

\subsection{Human Evaluation}
To complement our automated evaluations and validate the MLLM-as-a-Judge, we conducted a formal human evaluation study. The primary goal is to determine if the MLLM-as-a-Judge corresponds with the subjective preferences of human annotators.
We focus on the video generation tasks (LongText2Video, Image2Video, and Video2Video). We collected generated video results from both our UniVA system and the baseline models for each task. Annotators were asked to judge each video based on a set of criteria identical to MLLM.

\begin{figure}[h!]
  \centering
   \includegraphics[width=1\linewidth]{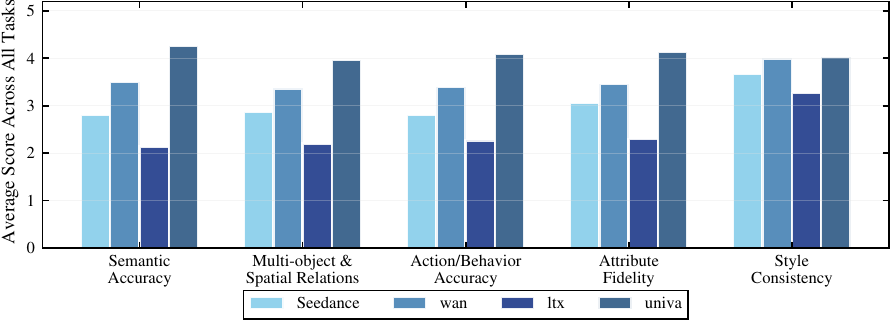}
   \caption{
   Results from the human evaluation study on video generation tasks.
   }
\vspace{-2mm}
   \label{fig:human-eval}
\end{figure}

In Figure \ref{fig:human-eval}, UniVA (darkest blue bar) emerges as the clear leader, achieving the highest human preference scores in four out of the five evaluated dimensions. This strong human preference aligns with the patterns observed in our automated metrics, confirming that our MLLM judge is a reliable proxy for genuine human perception.

\subsection{Qualitative Case Studies}

\begin{figure}[h!]
  \centering
   \includegraphics[width=1\linewidth]{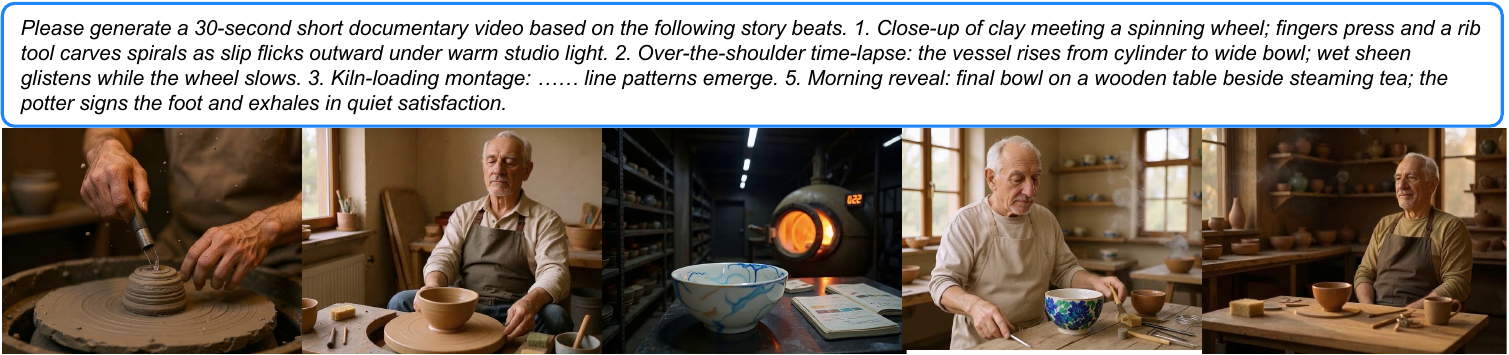}
   \caption{
   UniVA accurately generates the sequential process of pottery making, demonstrating strong temporal consistency and object persistence as the bowl evolves from clay to a finished product.
   }
   \label{fig:cs1}
\end{figure}

\begin{figure}[h!]
  \centering
   \includegraphics[width=1\linewidth]{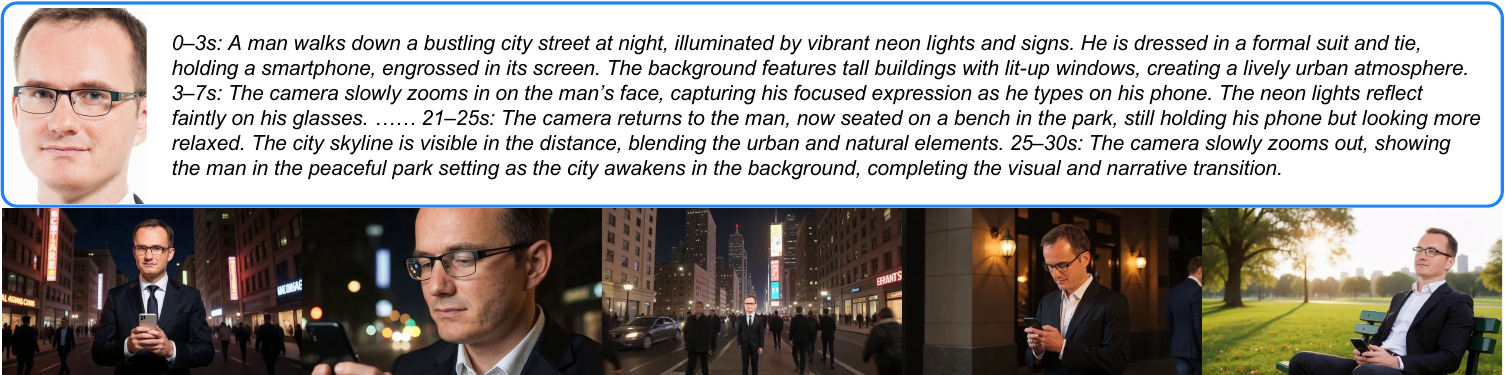}
   \caption{
   UniVA maintains the protagonist's identity flawlessly across drastically different scenes, lighting conditions (night vs. day), and camera angles, showcasing its advanced capability for robust, long-form character preservation.
   }
   \label{fig:cs2}
\end{figure}

\begin{figure}[h!]
  \centering
   \includegraphics[width=1\linewidth]{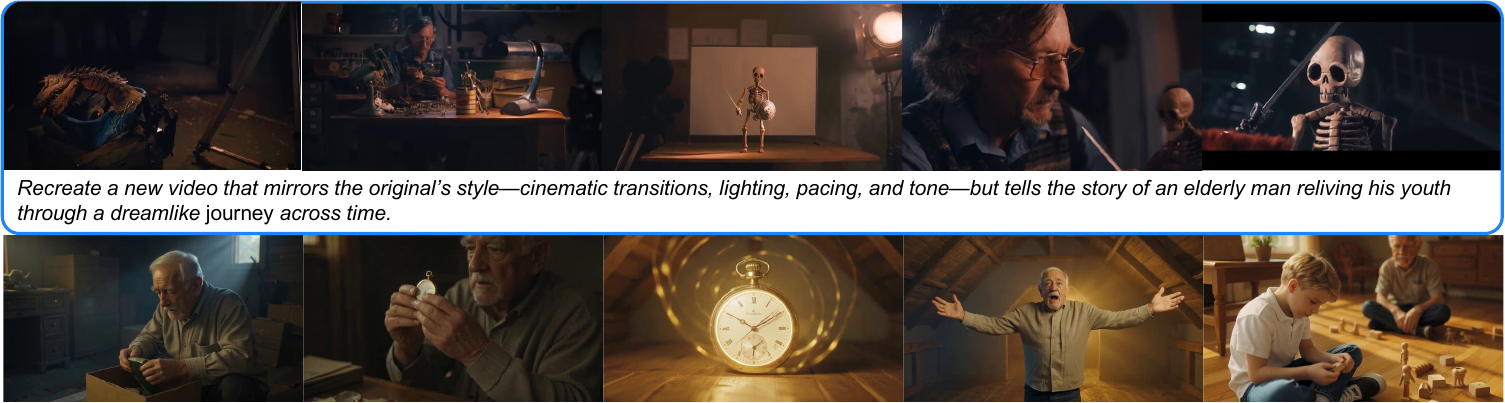}
   \caption{
   UniVA interprets an abstract prompt to generate a complex narrative. It orchestrates a non-linear story arc, proving its capability as an intelligent storyteller powered by sophisticated planning.
   }
\vspace{-2mm}
   \label{fig:cs3}
\end{figure}

\begin{figure}[h!]
  \centering
   \includegraphics[width=1\linewidth]{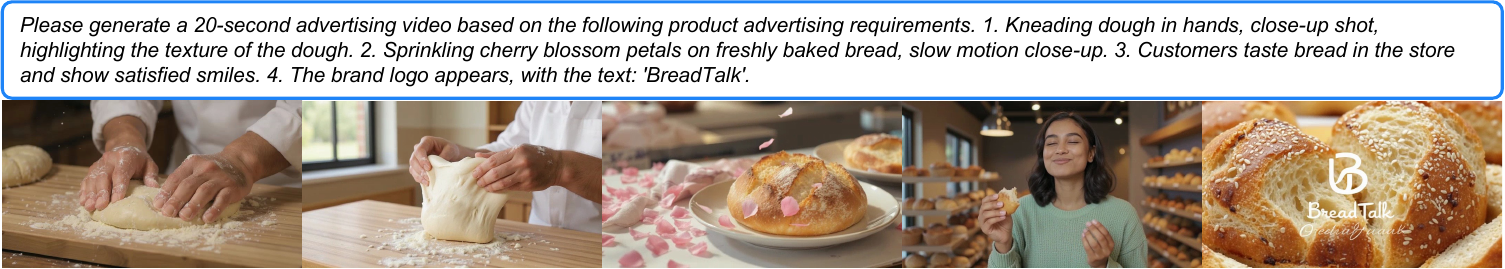}
   \caption{
   UniVA generates a coherent 20-second commercial that accurately follows the structured sequence of requirements—from kneading dough and showing customer reactions to applying the final brand logo.
   }
   \label{fig:cs4}
\end{figure}

\begin{figure}[h!]
  \centering
   \includegraphics[width=1\linewidth]{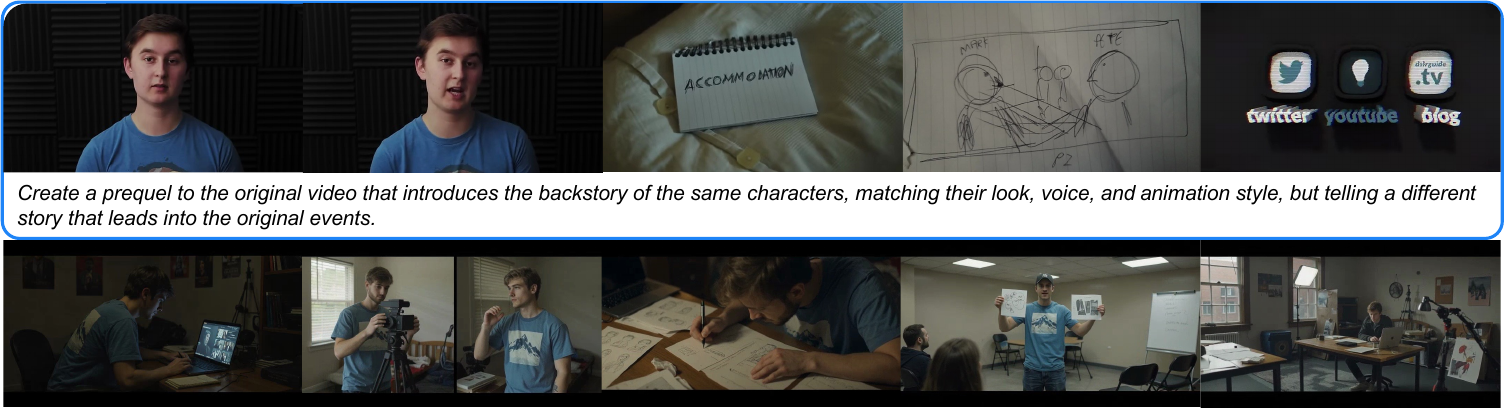}
   \caption{
   Given an original video, the agent not only maintains the original characters' style but also logically constructs a new backstory.
   }
   \label{fig:cs5}
\end{figure}

\begin{figure}[h!]
  \centering
   \includegraphics[width=1\linewidth]{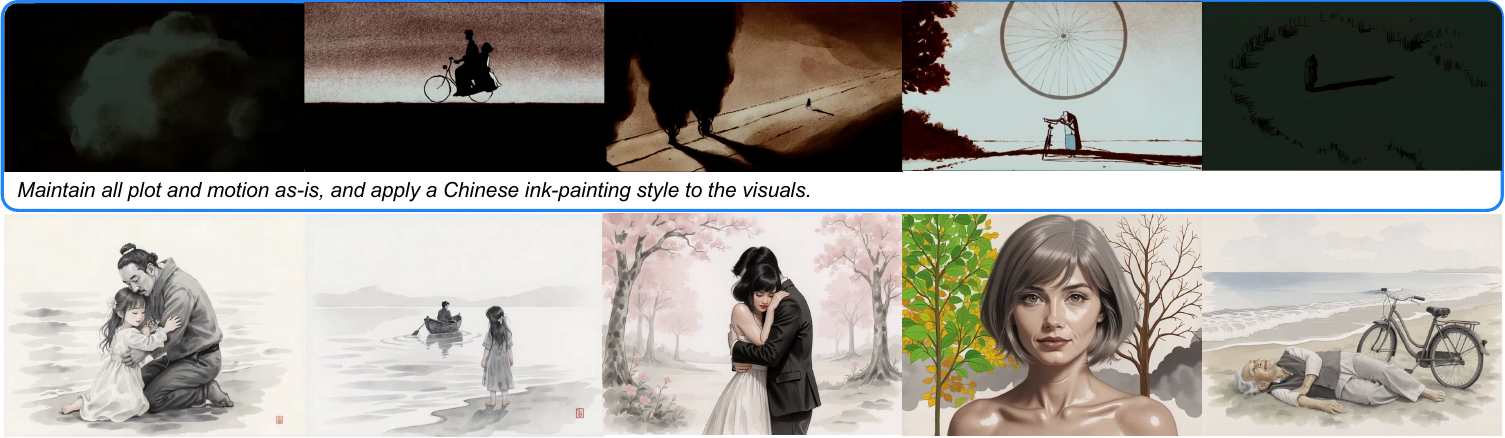}
   \caption{
   UniVA successfully applies a ``Chinese ink-painting style'' to the visuals while precisely maintaining the original video's plot, character motion, and scene composition.
   }
   \label{fig:cs6}
\end{figure}

\begin{figure}[h!]
  \centering
   \includegraphics[width=1\linewidth]{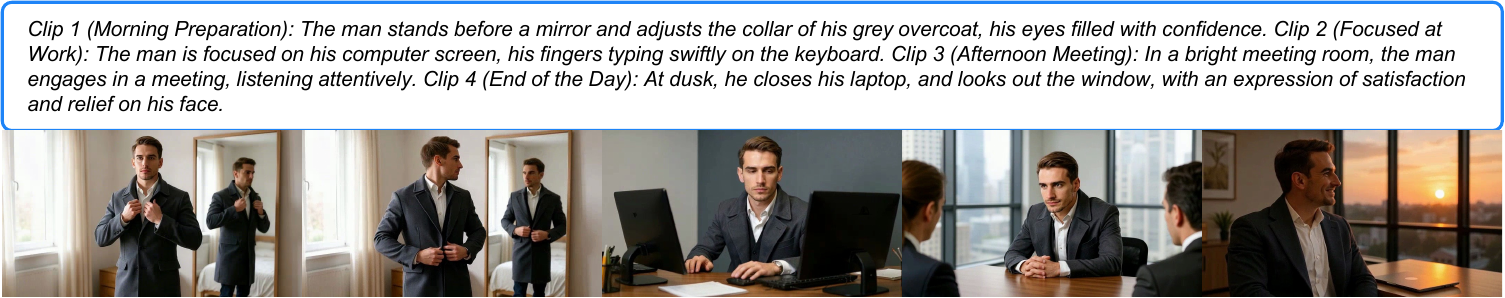}
   \caption{
   UniVA can well follow the user's long instructions and ensure consistency of characters in long videos.
   }
   \label{fig:cs7}
\end{figure}

\begin{figure}[h!]
  \centering
   \includegraphics[width=1\linewidth]{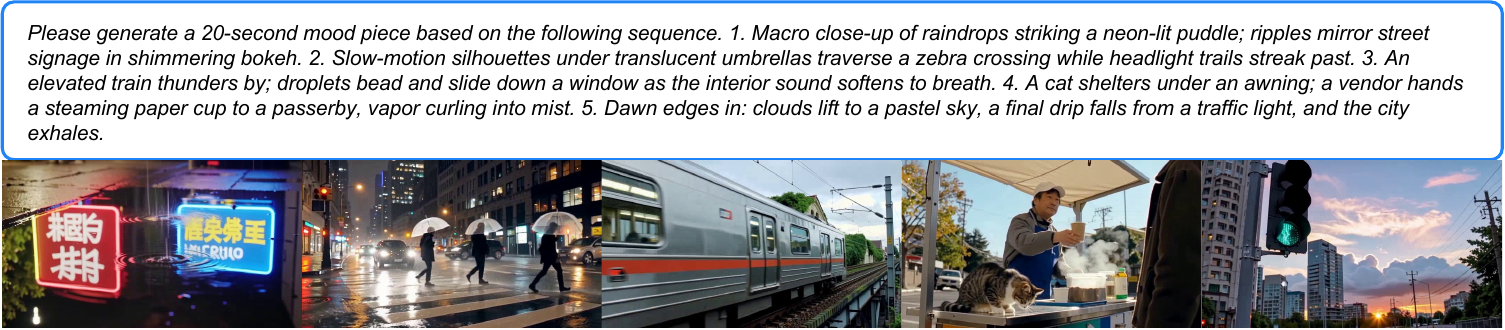}
   \caption{
   UniVA can understand and generate complex multi-camera scene transitions, producing multi-camera long videos.
   }
   \label{fig:cs8}
\end{figure}

\begin{figure}[h!]
  \centering
   \includegraphics[width=1\linewidth]{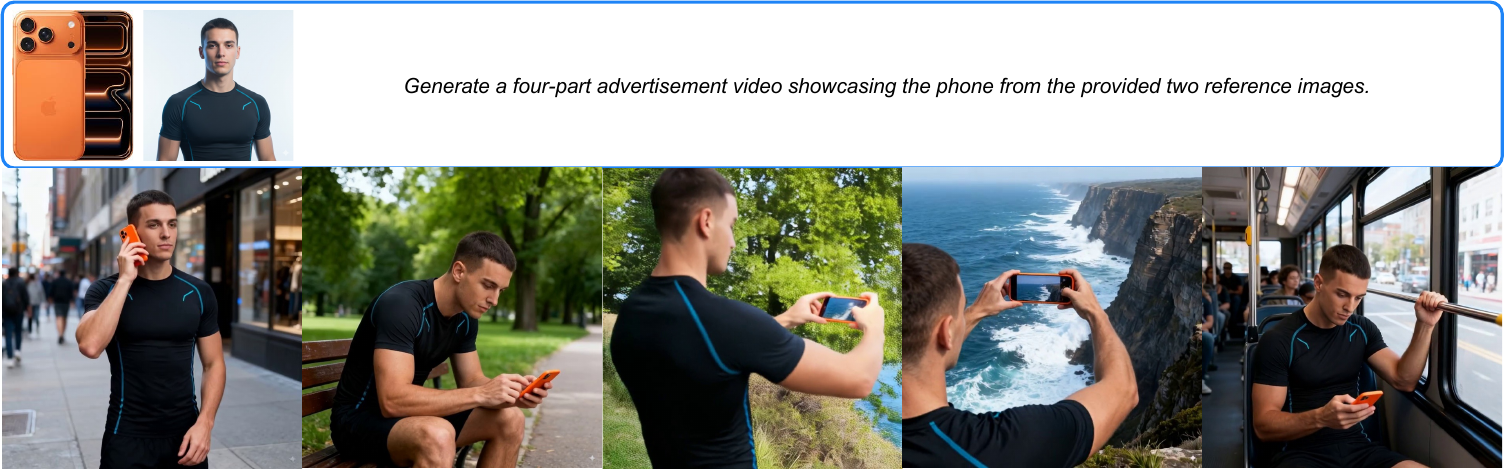}
   \caption{
   Univa can also maintain consistency well for multiple entity references.
   }
   \label{fig:cs9}
\end{figure}

\begin{figure}[h!]
  \centering
   \includegraphics[width=1\linewidth]{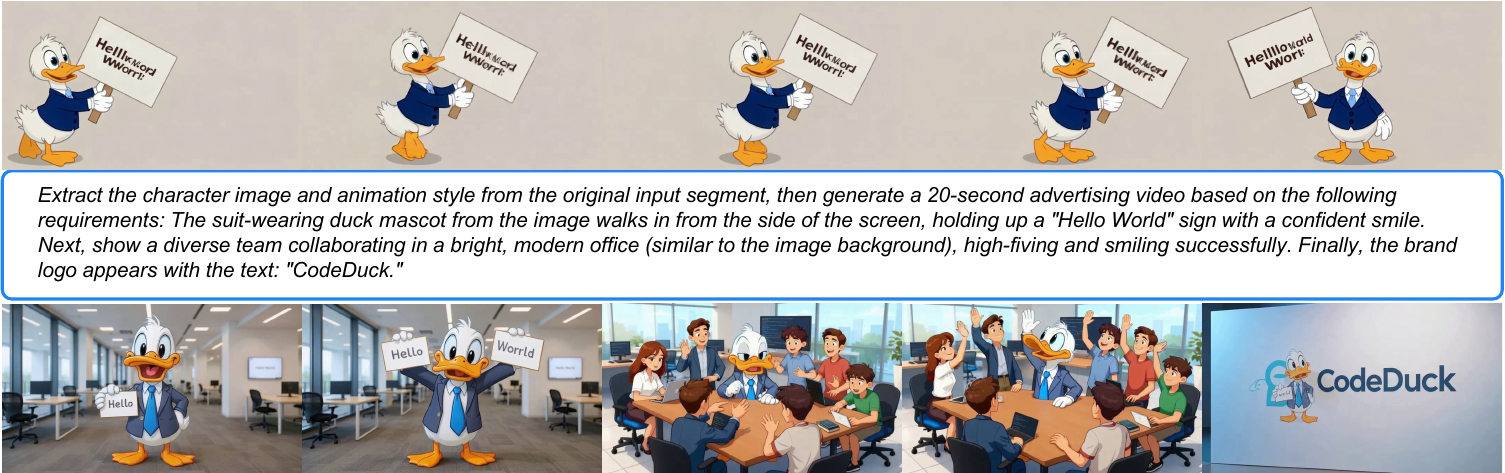}
   \caption{
   Univa can accurately analyze and understand the characters and style of a video, then seamlessly apply them to generate content.
   }
   \label{fig:cs10}
\end{figure}

\begin{figure}[h!]
  \centering
   \includegraphics[width=1\linewidth]{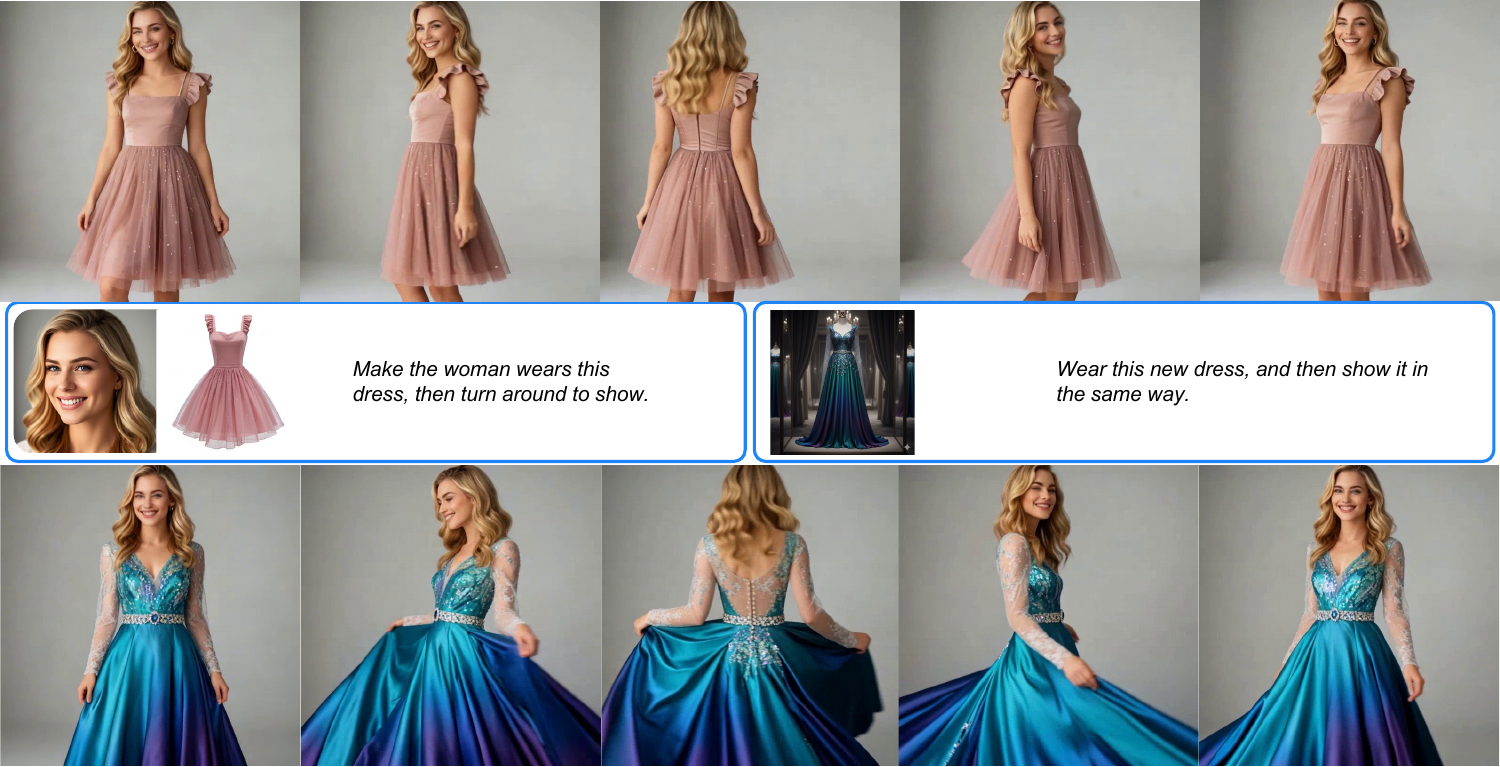}
   \caption{
   Univa can perform tasks through multi-turn dialogues by leveraging memory mechanisms and context.
   }
   \label{fig:cs12}
\end{figure}

To provide a more intuitive understanding of these quantitative results, we present a series of qualitative case studies (Figures~\ref{fig:cs1}–\ref{fig:cs12}).

UniVA delivers a highly automated, \textbf{proactive}, and \textbf{interactive} creation experience. It not only iterates on stories through \textbf{multi-round co-creation} and \textbf{deep memory context} (e.g., the outfit change in Figure~\ref{fig:cs12}) but also \textbf{proactively plans} steps, understanding implicit user intent and suggesting optimizations. Concurrently, as an \textbf{industrial-grade universal video fabric}, UniVA demonstrates its powerful extensibility: it can handle \textbf{any-conditioned} inputs, such as analyzing characters and styles from a video (Figure~\ref{fig:cs3}) or maintaining multi-entity references from images (Figure~\ref{fig:cs9}). And it can manage \textbf{complex narratives}, such as precisely following long instructions (Figure~\ref{fig:cs8}) and orchestrating multi-camera scenes (Figure~\ref{fig:cs9}), achieving end-to-end, professional-grade production from understanding and editing to generation.


For more demo videos or direct use experience, please visit: \url{http://univa.online/}.

\section{Conclusion}

In this work, we introduce UniVA, a unified agentic framework designed to tackle the next frontier of video intelligence. 
We argue that progress in video domain requires a paradigm shift from developing isolated, single-task models to creating integrated systems capable of complex, collaborative workflows. 
To this end, our primary contributions are the development of the powerful and extensible UniVA platform, the demonstration of its emergent synergistic capabilities as an omnipotent video generalist, and the release of UniVA-Bench to rigorously measure such advancements.
Our experiments validate UniVA's breadth, demonstrating competitive performance across a wide array of video tasks. 
More profoundly, we reveal its depth: through ``Agentic Synergy'', enabled by the dynamic management of information flow between tools, UniVA solves complex consistency problems intractable for siloed models. 
This confirms that UniVA is not merely a collection of tools, but an engine for generating emergent intelligence.
We hope that UniVA and UniVA-Bench will inspire future video intelligence research into this new generation of integrated, synergistic AI systems.

\bibliographystyle{assets/plainnat}
\bibliography{paper}

\clearpage
\newpage
\beginappendix

\input{appendix}

\end{document}

%% file: math_commands.tex
\usepackage{amsmath,amsfonts,bm}

\def\eqref#1{equation~\ref{#1}}

\def\1{\bm{1}}

\DeclareMathAlphabet{\mathsfit}{\encodingdefault}{\sfdefault}{m}{sl}
\SetMathAlphabet{\mathsfit}{bold}{\encodingdefault}{\sfdefault}{bx}{n}



%% file: abstract.tex
\abstract{
While specialized AI models excel at isolated video tasks like generation or understanding, real-world applications demand complex, iterative workflows that combine these capabilities. 
To bridge this gap, we introduce \textbf{UniVA}, an \textit{open-source, omni-capable multi-agent framework} for next-generation video generalists that unifies video understanding, segmentation, editing, and generation into cohesive workflows. 
UniVA employs a \textit{Plan-and-Act dual-agent architecture} that drives a \textit{highly automated and proactive workflow}: a planner agent interprets user intentions and decomposes them into structured video-processing steps, while executor agents execute these through modular, MCP-based tool servers (for analysis, generation, editing, tracking, \textit{etc.}). 
Through a \textit{hierarchical multi-level memory} (global knowledge, task context, and user-specific preferences), UniVA sustains long-horizon reasoning, contextual continuity, and inter-agent communication, enabling \textit{interactive and self-reflective video creation} with full traceability. 
This design enables \textit{iterative and any-conditioned} video workflows (\textit{e.g.}, text/image/video-conditioned generation $\rightarrow$ multi-round editing $\rightarrow$ object segmentation $\rightarrow$ compositional synthesis) that were previously cumbersome to achieve with single-purpose models or monolithic video-language models. 
We also introduce \textbf{UniVA-Bench}, a benchmark suite of multi-step video tasks spanning understanding, editing, segmentation, and generation, to rigorously evaluate such agentic video systems. 
Both UniVA and UniVA-Bench are fully open-sourced, aiming to catalyze research on \textit{interactive, agentic, and general-purpose video intelligence} for the next generation of multimodal AI systems.
}

%% file: appendix.tex
This supplementary material includes the following sections:
\begin{itemize}
    \item Detailed Methodology (cf. \S\ref{sec:Detailed_Methodology})
    \item UniVA-Bench (cf. \S\ref{sec:unibench_app})
    \item Detailed Experiment Settings (cf. \S\ref{sec:detailed_settings})
\end{itemize}

\section{Detailed Methodology}
\label{sec:Detailed_Methodology}
Since UniVA integrates multiple functionalities within a large-scale system, it is essential to clarify its design philosophy.
In this section, we present the guiding principles and functional workflow that underpin the framework, providing a comprehensive view of our design.

\subsection{Principles}
Here we provide a set of design principles that guide the construction of our UniVA framework. 

\paragraph{1. Unified \& Modular Architecture.}
A comprehensive and extensible system requires a modular architecture.
In UniVA, all capabilities—from SoTA generation models to simple non-AI tools—are integrated as decoupled functional modules.
These modules are invoked via a unified MCP, allowing them to be updated or replaced in a plug-and-play fashion. 
This principle is the foundation for the system's industrial-level production capabilities and ensures it can consistently deliver cinematic-quality output by leveraging the best available tools.

\paragraph{2. Separation of Plan \& Act for Complex Workflows.}
At the core of the agent's operation is a strict Plan-Act separation, which realizes the dual agent architecture described previously.
A Planner agent interprets high-level user intent and decomposes it into a logical sequence of steps.
An Executor agent then carries out each step by invoking the appropriate tools. This separation is crucial for managing long-horizon tasks and allows the system to robustly handle complex, multi-step video production pipelines.

\paragraph{3. Proactive, Goal-Oriented Autonomy.}
Crucially, the Planner is more than a passive task decomposer; it is designed for a high degree of automation and proactive behavior.
The agent actively evaluates intermediate results against the inferred user goal.
If an output does not align with the objective—as shown in the teaser, where the agent decides a video is insufficient—it initiates self-reflection to flexibly adapt its plan.
This ability to autonomously course-correct is the key to accomplishing an entire production pipeline from a single user query.

\paragraph{4. Hierarchical Memory for Immersive Interaction.}
The framework's multi-level memory mechanism is what enables iterative, multi-round interactions and deeply immersive creative experiences.
This hierarchy consists of global memory for persistent knowledge, task-specific memory to maintain context for the current workflow, and user memory to track preferences.
This design ensures contextual continuity, allowing users to refine and build upon their creations over extended interactions.

\paragraph{5. Composition of Atomic Operations into Robust Workflows.}
To effectively handle the composite and iterative workflows mentioned earlier, the framework strikes a balance between flexibility and reliability.
All complex functionalities are built upon a set of fine-grained atomic operations.
The Planner can creatively combine these atomic operations to solve novel problems.
For common, high-stakes tasks, these operations are organized into pre-defined workflow patterns to ensure robust and predictable execution.
This dual approach provides the system with both the versatility for creative exploration and the stability required for industrial-grade production.

Figure~\ref{fig:process} showcases how UniVA’s components work in synergy, revealing both its depth in handling complex, autonomous tasks and its breadth in supporting interactive, multi-tool creation.

The one-prompt task (left panel) exemplifies the system's depth. Faced with a complex command, the \emph{Plan-Act agent} autonomously decomposes the goal and orchestrates a sequence of tools via the \emph{MCP Servers}. By managing the information flow through the \emph{Memory Mechanism}, it effectively connects different capabilities, such as using an \emph{understanding} tool to empower a \emph{generation} tool. This enables the agent to collaboratively use multiple tools to achieve a sophisticated goal in a single pass.

Conversely, the multi-round task (right panel) highlights the system's breadth. It provides a powerful platform with a wide array of tools that users can flexibly combine through iterative interaction. Each command triggers a new Plan-Act cycle, where the agent leverages context from the \emph{Memory Mechanism} (e.g., a segmentation mask) to execute the next step. This demonstrates how our architecture supports a flexible, stateful, and collaborative creative process.

\begin{figure*}[!t]
  \centering
\includegraphics[width=0.94\textwidth]{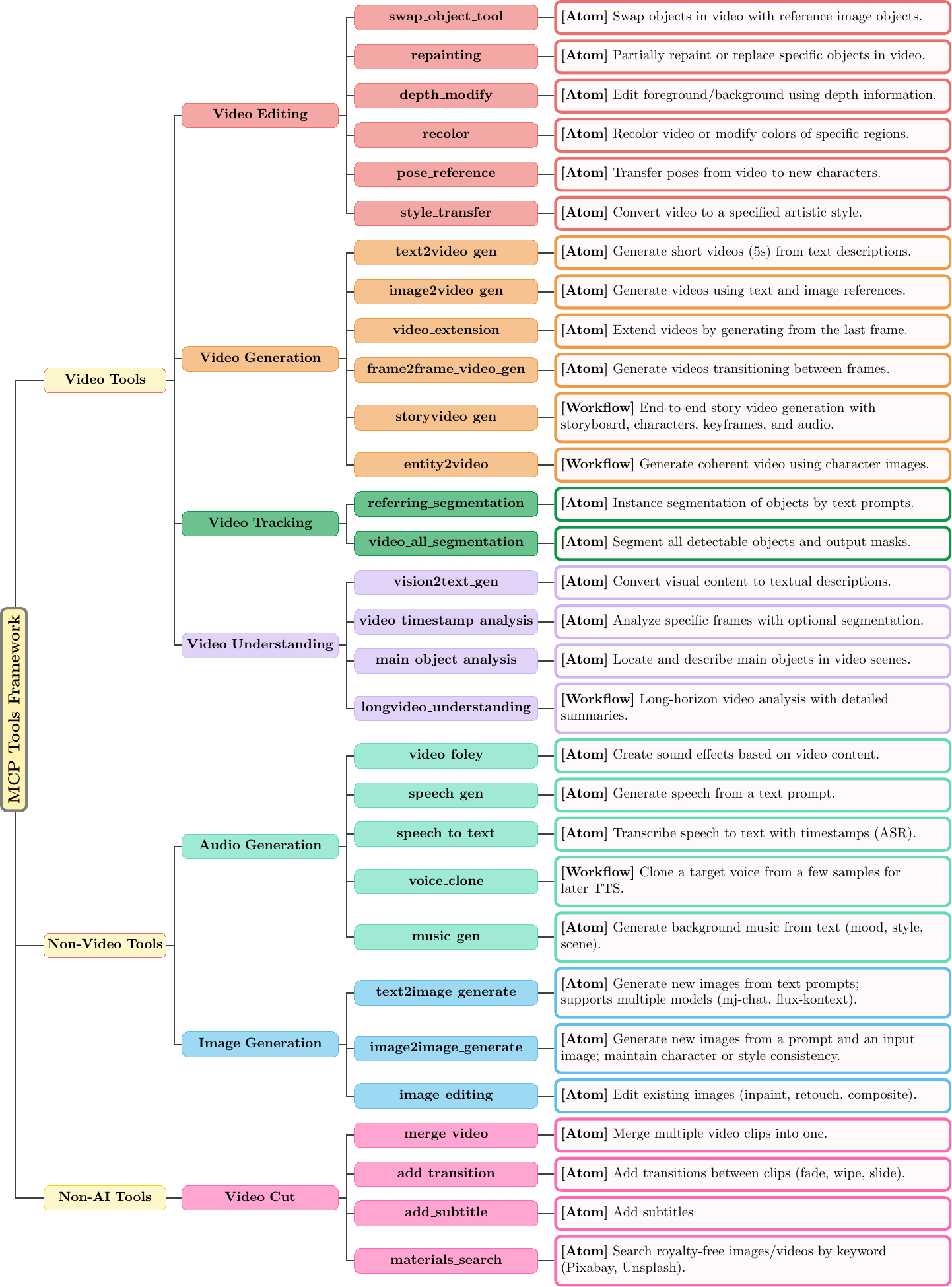}
  \caption{A three-level taxonomy of MCP tools: modules (level-1), tools (level-2), and leaf boxes summarizing name, type, and functionality.}
  \label{fig:org}
\end{figure*}

\subsection{MCP Function Walkthrough}
\label{sec:functions walkthrough}

UniVA is equipped with an extensive, modular toolset integrated via the MCP. This ``plug-and-play'' architecture enables the agent framework to invoke a diverse range of specialized functions.
As shown in Figure \ref{fig:org}, these tools are organized into three main categories: Video Tools, Non-Video Tools, and Non-AI Tools.

Each function is classified as either an \texttt{[Atom]} or a \texttt{[Workflow]}:
\begin{itemize}
    \item \textbf{\texttt{[Atom]}}: A fundamental, single-purpose operation, such as generating an image from text.
    \item \textbf{\texttt{[Workflow]}}: A higher-level function that composes multiple atomic tools to complete a multi-step task, such as generating an entire story video.
\end{itemize}

\subsubsection{Video Tools}

This core category encompasses functionalities for video creation, modification, and analysis.

\paragraph{Video Editing.} 
We use VACE\footnote{\url{https://github.com/ali-vilab/VACE}} \citep{jiang2025vaceallinonevideocreation} and Runway-Gen4-aleph\footnote{\url{https://runwayml.com/research/introducing-runway-aleph}} to provide granular control over the visual elements within a video.
\begin{itemize}
    \item \textbf{\texttt{swap\_object\_tool}} \texttt{[Atom]}: Swaps objects in a video with those from a reference image.
    \item \textbf{\texttt{repainting}} \texttt{[Atom]}: Repaints or replaces specific objects within a video.
    \item \textbf{\texttt{depth\_modify}} \texttt{[Atom]}: Edits the foreground or background of a video using depth information.
    \item \textbf{\texttt{recolor}} \texttt{[Atom]}: Recolors an entire video or modifies the colors of specific regions.
    \item \textbf{\texttt{pose\_reference}} \texttt{[Atom]}: Transfers poses and movements from a source video character to a new one.
    \item \textbf{\texttt{style\_transfer}} \texttt{[Atom]}: Applies a specified artistic style to a video.
\end{itemize}

\paragraph{Video Generation.} We use Seedance\footnote{\url{https://seed.bytedance.com/en/seedance}} \citep{gao2025seedance10exploringboundaries} to creates new video content from various inputs.
\begin{itemize}
    \item \textbf{\texttt{text2video\_gen}} \texttt{[Atom]}: Generates short videos (approx. 5s) from text descriptions.
    \item \textbf{\texttt{image2video\_gen}} \texttt{[Atom]}: Generates videos from a text prompt and an image reference.
    \item \textbf{\texttt{video\_extension}} \texttt{[Atom]}: Extends a video by generating subsequent frames.
    \item \textbf{\texttt{frame2frame\_video\_gen}} \texttt{[Atom]}: Generates a video transitioning between a start and end frame.
    \item \textbf{\texttt{storyvideo\_gen}} \texttt{[Workflow]}: End-to-end story video generation, including storyboard, characters, keyframes, and audio.
    \item \textbf{\texttt{entity2video}} \texttt{[Workflow]}: Generates a coherent video using a set of character images.
\end{itemize}

\paragraph{Video Tracking.} We use SAM2\footnote{\url{https://github.com/facebookresearch/sam2}} \citep{ravisam} and Sa2VA\footnote{\url{https://github.com/bytedance/Sa2VA}} \citep{yuan2025sa2va} to identify and isolate objects or regions within a video.
\begin{itemize}
    \item \textbf{\texttt{referring\_segmentation}} \texttt{[Atom]}: Segments video objects based on text prompts.
    \item \textbf{\texttt{video\_all\_segmentation}} \texttt{[Atom]}: Segments all detectable objects in a video and outputs their masks.
\end{itemize}

\paragraph{Video Understanding.} We use InternVL3\footnote{\url{https://github.com/OpenGVLab/InternVL}} \citep{zhu2025internvl3exploringadvancedtraining} to analyze and extract semantic information from video.
\begin{itemize}
    \item \textbf{\texttt{vision2text\_gen}} \texttt{[Atom]}: Generates a textual description of a video's visual content.
    \item \textbf{\texttt{video\_timestamp\_analysis}} \texttt{[Atom]}: Analyzes specific frames, with optional segmentation for focused analysis.
    \item \textbf{\texttt{main\_object\_analysis}} \texttt{[Atom]}: Locates and describes the main objects in video scenes.
    \item \textbf{\texttt{longvideo\_understanding}} \texttt{[Workflow]}: Analyzes long videos to provide detailed summaries and insights.
\end{itemize}

\subsubsection{Non-Video Tools}
This category includes functionalities for audio and image creation, editing, and synchronization.

\paragraph{Audio Generation.} We use MiniMax-Speech\footnote{\url{https://www.minimax.io/audio}} \citep{zhang2025minimaxspeechintrinsiczeroshottexttospeech} and ThinkSound\footnote{\url{https://github.com/FunAudioLLM/ThinkSound}} \citep{liu2025thinksoundchainofthoughtreasoningmultimodal} to create highly realistic sound effects and voice acting.
\begin{itemize}
    \item \textbf{\texttt{video\_foley}} \texttt{[Atom]}: Create and sync sound effects (foley) to visual events.
    \item \textbf{\texttt{speech\_gen}} \texttt{[Atom]}: Generate speech from a text prompt.
    \item \textbf{\texttt{speech\_to\_text}} \texttt{[Atom]}: Transcribe speech to text with timestamps (ASR).
    \item \textbf{\texttt{voice\_clone}} \texttt{[Workflow]}: Clone a target voice from a few samples for later TTS.
    \item \textbf{\texttt{music\_gen}} \texttt{[Atom]}: Generate background music from text (mood, style, scene).
\end{itemize}

\paragraph{Image Generation.} We use FLUX-Kontext\footnote{\url{https://github.com/black-forest-labs/flux}} \citep{labs2025flux1kontextflowmatching} generate high-quality images and edit images.
\begin{itemize}
    \item \textbf{\texttt{text2image\_generate}} \texttt{[Atom]}: Generate images from text prompts.
    \item \textbf{\texttt{image2image\_generate}} \texttt{[Atom]}: Generate a new image from a prompt conditioned on an input image for style/identity consistency.
    \item \textbf{\texttt{image\_editing}} \texttt{[Atom]}: Edit existing images (inpaint, retouch, composite).
\end{itemize}

\subsubsection{Non-AI Tools}
This category provides deterministic utilities for cutting, merging, and augmenting video materials.

\paragraph{Video Cut.}
\begin{itemize}
    \item \textbf{\texttt{merge\_video}} \texttt{[Atom]}: Merge multiple clips into a single sequence.
    \item \textbf{\texttt{add\_transition}} \texttt{[Atom]}: Add transitions between clips (fade, wipe, slide).
    \item \textbf{\texttt{add\_subtitle}} \texttt{[Atom]}: Add subtitles.
    \item \textbf{\texttt{materials\_search}} \texttt{[Atom]}: Search royalty-free images/videos by keyword (e.g., Pixabay, Unsplash).
\end{itemize}

\textbf{Remark.}
We note that everything in the current version of \textsc{UniVA} is inherently extensible, 
benefiting from the modular design of MCP. 
Each function—whether a low-level atomic operation or a high-level workflow, i.e., can be seamlessly registered, replaced, or expanded without modifying the core architecture. 
This flexibility allows the system to incorporate newly emerging video models, APIs, or custom tools in a plug-and-play manner. 
In essence, \textsc{UniVA} provides an open and evolving ecosystem rather than a closed pipeline, making it capable of continuously growing towards a truly universal and omni-capable video agent.

\subsection{Core Prompt}
We will open-source all code resources of UniVA, including all prompts used in the agents. Below are some core prompts utilized in the agents. To view all prompts, please refer to the code.

\begin{tcolorbox}[colback=gray!10, colframe=gray!50, title={Planner Prompt}, breakable]
\input{prompts/plan_prompt}
\end{tcolorbox}

\begin{tcolorbox}[colback=gray!10, colframe=gray!50, title={Executor Prompt}, breakable]
\input{prompts/act_prompt}
\end{tcolorbox}

\begin{tcolorbox}[colback=gray!10, colframe=gray!50, title={Storyboard Prompt}, breakable]
\input{prompts/storyboard}
\end{tcolorbox}

\section{Extended Description on UniVA-Bench}
\label{sec:unibench_app}

\subsection{Data Curation}
\label{sec:curation}
\paragraph{Understanding (Long-Video QA).}
We randomly sampled 10 videos from Video-MME \citep{fu2025video} and used Gemini 2.5 Pro to generate multiple-choice QA pairs based on the perspectives shown in Table~\ref{table:data_und}. The task specifies that each video corresponds to 10 questions, and all answers must be provided within a single inference.

\paragraph{Generation.}
In the data curation stage, for the LongText2Video task, we first use GPT to generate a clear storyboard, then rewrite it into long and noisy prompts. For the Image/Entities2Video task, we first sample 10 data points from Opens2v-nexus \citep{yuan2025opens2v}. We then rewrite the original prompts into longer and noisier versions.

For the Video2Video task, in order to better approximate real-world scenarios, we divide it into three settings:
\begin{itemize}
    \item \textit{Story alignment:} Given a video, modify its style according to the prompt while keeping all other aspects unchanged.

    \item \textit{Style alignment:} Given a video, modify the storyline according to the prompt while preserving the original video’s style, characters, and semantics.

    \item \textit{Semantic alignment:} Given a video, modify both the style and storyline according to the prompt while retaining the original characters and other semantic elements (e.g., generating a sequel to the video). For each task, we sampled 10 videos from SF20k \citep{ghermi2024long}, then manually generate prompts for each video.

\end{itemize}

\begin{table}[h!]
\setlength{\tabcolsep}{8mm}
\centering
\caption{Key dimensions for analyzing video shots and editing.}
\begin{tabular}{ll}
\toprule
\textbf{Category} & \textbf{Dimension} \\ \midrule
\multirow{6}{*}{Intra-frame} 
  & 1. Shot Size \\
  & 2. Shot Angle \\
  & 3. Shot Location \\
  & 4. Shot Subject \\
  & 5. Shot Type (composition) \\
  & 6. Shot Color (grading/tonality) \\ \midrule
\multirow{2}{*}{Intra-shot} 
  & 7. Shot Motion (camera movement) \\
  & 8. Shot Speed \\ \midrule
\multirow{2}{*}{Inter-shot} 
  & 9. Cut Type \\
  & 10. Transition \\ \bottomrule
\end{tabular}
\label{table:data_und}
\end{table}

\paragraph{Editing (Long Video).}
We sampled 10 videos from SF20k \citep{ghermi2024long}, then manually curated the editing prompt based on the content of the video.

\paragraph{Segmentation (Long Video).}
We randomly concatenated clips from DAVIS2017 \citep{perazzi2016benchmark}, resulting in 10 segmentation task instances that involve occlusions and cover diverse scenes.

\paragraph{Agentic probing sets.}
We include
(1) a 50-instance storyboard$\rightarrow$user-intent planning set to compare Single-Agent vs. Plan-Act, and
(2) a set of standard pipeline tasks with expert references to assess wPED, DepCov, and ReplanQ under injected failures.
Memory analyses consider \emph{trace memory} (historical trajectories), \emph{user memory} (preferences), and \emph{task memory} (e.g., storyboards).

\subsection{Evaluation Suite}
\label{sec:metrics}

\subsubsection{Subject metrics (task quality).}
  \paragraph{CLIP Score (command following).}
  Measures text-video alignment between the user instruction (or storyboard-derived captions) and generated/edited outputs. We report the average CLIP similarity over sampled frames/clips; higher is better.
  
  \paragraph{DINO Score (subject consistency).}
  Measures referential/identity stability by comparing DINO features between reference entities (images/key frames) and generated/edited frames; the higher the better.
  
  \paragraph{Segmentation: J/F/mIoU.}
  We report region (\textbf{J}-mean, IoU) and boundary (\textbf{F}-mean) quality, as well as \textbf{J\&F}-mean; higher is better.
  
  \paragraph{Understanding Score.}
  Normalized accuracy over curated long-video QA pairs that span both semantics and aesthetics (shot transitions, style, narrative).

\subsubsection{MLLM as a judge (preference).}
To complement subject metrics, we perform pairwise preference judgments using an open-source judge (e.g., \textbf{InternVL-3-78B} \citep{zhu2025internvl3exploringadvancedtraining}) and a closed-source judge (e.g., \textbf{Gemini-2.5-pro} \citep{comanici2025gemini25pushingfrontier}).
Judges are provided with the instructions, any relevant references, and debiased captions; preferences are aggregated via majority voting, with ties being discarded.
We report average preference rates and include significance tests when applicable.

To ensure consistent and unbiased evaluation, we used a standardized prompt template for our MLLM judge. The template was designed to be comprehensive and force a structured output.

\begin{tcolorbox}[colback=gray!10, colframe=gray!50, title={MLLM Prompt}, breakable]
\input{prompts/mllm_prompt}

\end{tcolorbox}

\subsubsection{Agentic metrics (planning \& recovery).}
To quantitatively evaluate the agent's planning capabilities, we designed three specialized metrics: Weighted Plan Edit Distance (wPED), Dependency Coverage (DepCov), and Re-planning Quality (ReplanQ). The precise definitions and calculation methods for these metrics are detailed below.

\paragraph{wPED (Weighted Plan Edit Distance).}
wPED measures the structural similarity between the sequence of tool names in an agent-generated plan ($P_{pred}$) and an expert-authored reference plan ($P_{ref}$). The score is derived from the classic Levenshtein edit distance, denoted as $L(A, B)$, which calculates the minimum number of single-item edits (insertions, deletions, or substitutions) needed to transform sequence A into sequence B.

The wPED score is calculated by normalizing this distance and inverting the result, ensuring that a higher score indicates a better alignment. The formula is:
\begin{equation}
\label{eq:wped}
\text{wPED} = 1 - \frac{L(P_{pred}, P_{ref})}{\max(\text{len}(P_{pred}), \text{len}(P_{ref}))}
\end{equation}
A higher wPED score (closer to 1.0) signifies a closer structural alignment to the expert plan.

\paragraph{DepCov (Dependency Coverage).}
DepCov evaluates the logical correctness of a generated plan ($P_{pred}$) by measuring its adherence to a set of fundamental, rule-based dependencies inherent to video production workflows. Our evaluation is based on a predefined set of rules, such as the principle that content generation must precede content editing.

Let $D(P_{pred})$ be the set of all dependency pairs $(u, v)$ identified in the plan $P_{pred}$ according to our rules, where tool $u$ must appear before tool $v$. Let $D_{sat}(P_{pred}) \subseteq D(P_{pred})$ be the subset of those pairs where this ordering is correctly satisfied. DepCov is then the fraction of satisfied dependencies:
\begin{equation}
\label{eq:depcov}
\text{DepCov} = \frac{|D_{sat}(P_{pred})|}{|D(P_{pred})|}
\end{equation}
A higher DepCov score indicates that the agent's plan is more logically sound and respects the procedural constraints of the task.

\paragraph{ReplanQ (Re-planning Quality).}
ReplanQ measures the agent's ability to efficiently and effectively recover from a simulated execution failure. The metric is designed to reward intelligent, minimal plan modifications.

Let $P_{orig}$ be the agent's initial plan, and let the failure occur at index $i$. The agent then generates a revised plan, $P_{replan}$. We compare the suffixes of both plans starting from the failure point, denoted as $P_{orig}[i:]$ and $P_{replan}[i:]$. ReplanQ is calculated using the same normalized Levenshtein distance as in wPED, applied only to these suffixes:
\begin{equation}
\label{eq:replanq}
\text{ReplanQ} = 1 - \frac{L(P_{orig}[i:], P_{replan}[i:])}{\max(\text{len}(P_{orig}[i:]), \text{len}(P_{replan}[i:]))}
\end{equation}
A higher ReplanQ score (closer to 1.0) indicates a more efficient and robust recovery, suggesting that fewer changes were required to correct the plan after the failure.

\section{Detailed Experiment Settings}
\label{sec:detailed_settings}

\subsection{UniVA's Configuration}

\begin{table}[h!]
\centering
\caption{UniVA’s Configuration}
\begin{tabular}{ll}
\hline
\textbf{Module} & \textbf{Model} \\ \hline
\textbf{Plan Agent} & Claude-sonnet-4\footnote{\url{https://www.anthropic.com/news/claude-4}} \\
\textbf{Act Agent} & Claude-sonnet-4 \\
\textbf{Video Generation Model} & Seedance-v4-480p \citep{gao2025seedance10exploringboundaries} \\
\textbf{Video Understanding Model} & InternVL3-38B \citep{zhu2025internvl3exploringadvancedtraining} \\
\textbf{Video Editing} & Runway Aleph \\
\textbf{Video Segmentation} & SAM-2 \citep{ravisam} and Sa2VA \citep{yuan2025sa2va} \\
\textbf{Image Generation Model} & FLUX-Kontext \citep{labs2025flux1kontextflowmatching} \\ \hline
\end{tabular}
\label{tab:modules}
\end{table}

\subsection{Baseline Configurations}
\paragraph{Generation.}
For all video generation tasks, we standardized the output resolution to 480p and a frame rate of 24 fps to ensure a fair comparison. 
For baselines that natively lacked support for multi-image or video-conditioned inputs, we implemented a standardized pre-processing pipeline to bridge the capability gap:
For the Entities2Video task, where some baselines only accept a single image, we first merged the multiple input reference images into a single composite image. This composite was then used as the input.
For the Video2Video task, for text-only baselines, we first employed a video captioning model (Qwen2.5-VL-72B \citep{bai2025qwen25vltechnicalreport}) to generate a detailed description of the source video. This generated caption was then prepended to the user's instruction prompt to guide the generation process.

The specific baseline models were configured as follows:
LTX-Video: We utilized the official model and followed the recommended settings provided in their public repository.
Seedance: We used the seedance-v1-pro-t2v-480p and seedance-v1-pro-i2v-480p from Wavespeed API\footnote{\url{https://wavespeed.ai/}}, consistent with our UniVA's generation module, to ensure a direct comparison of the agentic framework's contribution.
Wan: We used the wan-2.2/t2v-480p \citep{wan2025wanopenadvancedlargescale} and wan-2.2/i2v-480p also via the Wavespeed API.

\paragraph{Understanding.}
For all the understanding tasks, we are using a frame rate of 1 fps and a maximum of 128 frames.

\paragraph{Editing.}
For the video editing task, we use Runway Aleph as the baseline model. In the baseline pipeline, videos are clipped into 5-second clips and sent to the Aleph model with a task prompt. Then, the edited video clips are merged together for evaluation.

\paragraph{Segmentation.}
For the video segmentation task, we use Sa2Va-4B as the baseline model, we directly send the video into the baseline model together with the segmentation prompt.

%% file: prompts/plan_prompt.tex
\textbf{\color{iorange}[Role]}\\
You are Univideo, an expert video generation and processing planner. Your task is to analyze user requests and create detailed, step-by-step execution plans using available tools. You must break down complex tasks into manageable steps and select appropriate tools for each operation.\\

\textbf{\color{ired}\uppercase{I. Available Tools Overview}} \\
The following is a detailed description of the tool and suggested usage scenarios. The tools are divided to atom and workflow. Atomic functions are basic, independent functional modules, usually low-level operations, such as a step or function in video generation. These operations exist independently and can be used to build more complex processes. A defined workflow is a series of operations with clear steps and sequence. You only need to call the workflow interface without having to manually combine each step yourself.\\

\textbf{1. Video Generation Tools}:\\
\textbf{1.1 Atom Functions}:\\
- text2video\_gen: Generates a short video (approx. 5 seconds) from a text description. This tool is ideal for creating new video content based solely on a textual prompt. Need a text prompt, return a video.\\
- image2video\_gen: Generates a short video (approx. 5 seconds) using a text prompt and an input image as a visual reference, the generated video will start with this input image. This tool is useful for creating videos that maintain visual consistency with a provided image while following the text instruction. Need a text prompt and an image, return a video.\\
- video\_extension: Extends an existing video based on a text prompt and the last frame of the input video. This tool is suitable for seamlessly continuing a video's narrative or expanding its duration. Need a text prompt and a video, return a video.\\
- frame2frame\_video\_gen: Generates a short video (approx. 5 seconds) that transitions between a specified first frame and a last frame, guided by a text prompt. This tool is effective for creating dynamic action sequences or smooth transitions between two frames. Need a text prompt and two images, return a video.\\
\textbf{1.2 Workflow Functions}:\\
- storyvideo\_gen: Generates a story-based video from a text prompt by creating a storyboard, generating character images, creating keyframes, generating video segments, and merging them into a final video.\\
- entity2video: Generates a story-based video from a text prompt and a list of character images by creating a storyboard, using the provided images for characters, generating keyframes, creating video segments, and merging them into a final video.\\

\textbf{2 Video Editing Tools}:\\
\textbf{2.1 Atom Functions}:\\
- swap\_object\_tool: Swaps a specified object in a target video with the corresponding object from a reference image. This function identifies all instances of a given object class (e.g., "person", "car") in the input video and replaces them with the object provided in the reference image, guided by a textual prompt.\\
- depth\_modify: Based on a text prompt, use depth information to edit or replace the foreground or background of a video. This function is specifically designed for video editing tasks that require distinguishing between foreground and background. It is suitable for intelligent video editing, such as replacing the background or changing the foreground color, while leaving the other content unchanged.\\
- recolor: Recolorize a video or modify the color of specific areas based on a text prompt, allowing for overall stylistic recoloring.\\
- pose\_reference: Based on a text prompt, transfer the motions of a person in a video to a new character while preserving the original motion sequence. This function implements the pose transfer functionality, resulting in a new character performing the motions from the old video.\\
- style\_transfer: Based on a text prompt, converts a video into a specified artistic style, achieving style transfer. This function extracts edges and contours to generate a line drawing video. This process retains the core structure and dynamic information of the original video, but removes its original colors and textures, providing an ideal structural foundation for applying the new style. It then "renders" the line drawing video, generating a video with the same content and dynamics as the original video, but with a completely new visual style.\\
- repainting: Partially repaint or replace a specific object in a video, changing its appearance or transforming it into something entirely new based on a text prompt. This function implements video inpainting or object replacement by first calling the `label` parameter to identify and locate a specific object in the video (e.g., "cat," "car"). The script generates a precise dynamic mask for the identified object while preserving the original video. This mask marks the area to be edited. Next, it fills in the content described in `prompt`, modifying or completely replacing the original object and ensuring that the new content blends seamlessly with the rest of the video.\\

\textbf{3 Video Understanding Tools}:\\
\textbf{3.1 Atom Functions}:\\
- vision2text\_gen: Analyzes and describes the content of a video or image based on a given prompt, converting visual information into text. This tool is useful for understanding ambiguous or complex visual inputs, providing detailed textual descriptions of the content.\\
\textbf{3.2 Workflow Functions}:\\
- video\_timestamp\_analysis: Analyzes a specific timestamp (frame) in a video and generates a detailed text description. This function extracts a single frame at a specified time and optionally performs instance segmentation on the image. Then it feeds the processed (or original) image into a large multimodal language model (such as Qwen-VL) to generate descriptive captions. The analysis results, including the timestamp, generated description, and image path, are saved to a JSON file. This tool is ideal for performing in-depth and detailed analysis of a specific moment in a video.\\
- main\_object\_analysis: Analyzes the specified primary object in a video and generates descriptive text. This function works through a two-stage process: First, it uses video referring segmentation to locate and isolate the target object in the video based on the given text label, generating a segmented video containing only that object. Then, it feeds this segmented video into a multimodal large language model to generate a detailed analysis and description of the object. This tool is useful for extracting specific objects from complex video scenes and gaining a deep understanding of them.\\

\textbf{4 Video Tracking Tools}:\\
\textbf{4.1 Atom Functions}:\\
- video\_referring\_segmentation: Performs video instance segmentation to identify and outline specific objects within a video based on a textual prompt. This tool is useful for tasks requiring precise object localization and tracking in video content. Need text prompt and original video path, return the video path with segmentation result.\\
- video\_all\_segmentation: Segments all detectable objects within a video, providing a comprehensive mask for each identified object. This tool is useful for general object detection and mask generation across an entire video.\\

\textbf{5 Image Generation Tools}:\\
\textbf{5.1 Atom Functions}:\\
- text2image\_generate: Generates a new image based on a textual prompt. This tool is useful for creating visual content from scratch.\\
- image2image\_generate: Generates a new image based on a text prompt and an input image, while maintaining consistency with characters or styles from the original image. This tool is useful for modifying existing images or generating new ones with specific visual references.\\

\textbf{\color{igreen}\uppercase{II. Core Planning Logic \& Tool Selection Guidelines}}\\

\textbf{1. Basic Calling Logic}:\\
- If the user does not provide semantic information about the visual content, use `vision2text\_gen` to understand it.
- For materials that are not provided by the user or that you lack, you should try to use existing tools to generate the required materials first, and then perform subsequent steps.\\
- Style consistency requires careful reference management.\\

\textbf{\color{blue}\uppercase{III. Plan Output Format}}\\
Generate a clear, numbered list of steps. Each step should include:\\
- What the step accomplishes\\
- Which tool to use\\
- Key parameters or considerations\\
- Do not output any extra content, including comments, other than what is shown in the following example.\\
- Pay attention to the number limits of tool inputs and outputs. For example, you can only input one image at a time, and if there are two materials, you need to call it twice. And each call should be a separate step.\\
- If there are user-provided materials, please specify the path in input\_requirements.\\
- Because the act model can only capture information about a single step, you should provide as much detailed information as possible for each step of the plan.\\

Example format:
\begin{verbatim}
{
  "task_analysis": "Brief description of the identified task type and approach",
  "execution_plan": {
    "total_steps": 3,
    "steps": [
      {
        "step_number": 1,
        "action_description": "What this step accomplishes",
        "tool": {
          "name": "tool_name",
          "purpose": "What this achieves",
          "input_requirements": [
            "required material 1",
            "required material 2"
          ]
        },
        "dependencies": [
          "step numbers this depends on, empty array if none"
        ],
        "status": "success/failure/ongoing/pending",
        "output": "output from this step, empty string if none"
      },
      {
        "step_number": 2,
        "action_description": "Next action description",
        "tool": {
          "name": "tool_name",
          "purpose": "What this achieves",
          "input_requirements": [
            "output from 1"
          ]
        },
        "dependencies": [1],
        "status": "success/failure/ongoing/pending",
        "output": "output from this step, empty string if none"
      }
    ]
  }
}
\end{verbatim}
\quad\\

\textbf{\color{ipurple}\uppercase{IV. Planning Updates}}\\
- At the beginning, the first step should have a status of ongoing, and the unexecuted steps should have a status of pending.\\
- After each ongoing step is executed, it returns the result of the execution, and the status of the planning is updated based on this result.\\
- Only sequential execution of steps is allowed, and at each time, only one step can be in the ongoing state.\\
- If a step completes with a failure status, the planning should be dynamically adjusted.\\
- Each time a plan is output, one of the STEPs must be in the ongoing state so that the ACT model can find which STEP needs to be executed.\\
- Determine if the plan ended successfully, and if it did, no further updates to the PLAN are needed, a short summary will suffice.\\

Always provide clear, actionable steps with specific tool selections and parameter recommendations.

%% file: prompts/act_prompt.tex
\textbf{\color{iorange}[Role]}\\
You are an intelligent video task execution assistant (Video Act LLM), specialized in executing video generation, editing, and processing plans. You follow plans provided by the Plan Model while conducting intelligent thinking, calling, wait and feedback throughout the video production workflow.\\

\textbf{\color{ired}I. Tool Calling Protocol:} \\

\textbf{1. MCP Tool Execution Format:}\\
All tools must be called using the standardized MCP format:
\begin{verbatim}
<use_mcp_tool>
<server_name>[server_name]</server_name>
<tool_name>[tool_name]</tool_name>
<arguments>
{
  "parameter_name": "parameter_value",
  "another_parameter": "another_value"
}
</arguments>
</use_mcp_tool>
\end{verbatim}

\textbf{2. Tool Execution Workflow:}\\
- Think about the step requirements and tool selection\\
- Call the appropriate tool using MCP format\\

\textbf{3. Critical Execution Rules:}\\
- One tool per message: Never call multiple tools simultaneously\\
- Sequential execution: Each tool call must be informed by previous results\\
- Error handling: If a tool fails, analyze the error and adjust approach\\

\textbf{\color{igreen}II. Video-Specific Execution Framework:}\\

\textbf{1. Plan Reception and Video Task Understanding}\\
- Receive detailed video execution plans from the Plan Model\\
- Understand video generation constraints (~5 second segments)\\
- Identify visual consistency requirements across video segments\\
- Recognize dependencies between video clips and reference materials\\

\textbf{2. Video-Centric Step-by-Step Execution}\\
\textbf{2.1 Pre-Execution Analysis for Video Tasks:}\\
Before each specific step is executed, the Plan tells you which tool to call. Your "thinking" needs to address the question - for that tool, what parameters should be filled in to accomplish the goal of the step? For example:
\begin{verbatim}
<thinking>
- Specified tool: generate_clip
- Purpose of call: Generate a 5-second animation
- Parameters to check: duration: 5 seconds
  - Required parameters: duration: 5
  - style: "Modern Flat"
  - scene_description: "City street at night with flashing neon lights"
  - resolution: "1920x1080" (determined by program or quality standards)
- Boundary conditions.
  - Duration must be exactly 5 seconds
  - File format is mp4
</thinking>
\end{verbatim}
Do not call the tool at this stage, only list and confirm the required parameters.\\

\textbf{2.2 Tool Calls:}\\
Initiate a tool call strictly using the MCP format. For example, to generate a 5-second animation:
\begin{verbatim}
<use_mcp_tool>
<server_name>video_gen_server</server_name>
<tool_name>generate_clip</tool_name>
<arguments>
{
  "duration": 5,
  "style": ‘Modern Flat’,
  "scene_description": "City street at night with neon lights flashing"
}
</arguments>
</use_mcp_tool>
\end{verbatim}
After sending it, don't output anything more and end this STEP execution.

%% file: prompts/storyboard.tex
\textbf{\color{iorange}[Role]}\\
You are a professional storyboard creator who can take a single-sentence user input (a complete video concept) and automatically break it down into a sufficiently detailed Storyboard. The resulting Storyboard must contain enough information so that a downstream video-generation model can render the entire video directly from it.\\

\textbf{\color{ired}I. Task Description:} \\
\textbf{1. Input}\\
   - The user will supply exactly one short sentence summarizing the video’s storyline or theme (for example: “A little girl chases a glowing butterfly through a forest and eventually reaches a mysterious lake”).\\
   - Do not alter the user’s input; build a full Storyboard based solely on that sentence.\\

\textbf{2. Output}\\
   - Return strictly valid JSON containing exactly three top-level fields:\\
     1. \texttt{"characters"}: A list of every character (and any anthropomorphized prop, if needed) that appears in the video, each with a unique \texttt{"id"}.\\
     2. \texttt{"shots"}: An array where each element is one shot object. The array order defines the playback sequence, and adjacent shots must flow smoothly to tell a coherent story.\\
     3. \texttt{"style"}: A concise, clear description of the overall visual style for the entire Storyboard (e.g., “Dreamlike Cartoon Style” or “Realistic Watercolor Style”).\\

\textbf{3. "characters" Array Specification}\\
   - List every main character, supporting character, or anthropomorphized prop that will appear on-screen.\\
   - Each entry must be an object with these keys:\\
     - \texttt{"id"}: A unique string identifier, start with "char\_" and then with the index number (e.g., \texttt{"char\_1"}, \texttt{"char\_2"}, …).\\
     - \texttt{"name"}: The character’s in-story label (e.g., “little girl”, “glowing butterfly”, “forest spirit deer”).\\
     - \texttt{"description"}: A purely static appearance description—no actions, emotions, or behaviors. Include enough detail so that an artist or model could draw the character: approximate age or size, hairstyle or wing design, clothing style, key facial or body features (e.g., “eyes shining with curiosity”, “wings tipped with silver flecks”). Always refer to them as “this [character]” rather than using a proper name in the description.\\
   - Note: Each character should include just one main person or object, and no more than three supporting characters or objects.\\

\textbf{4. "shots" Array Specification}\\
   - The length of this array is determined automatically to cover the entire storyline so that you produce a complete video. For roughly a one-minute video, that usually means around 12–15 shots; for a three-minute video, around 30–40 shots. Adjust as needed so the story fits naturally.\\
   - Each shot object must include exactly these fields:\\
     1. \texttt{"id"}: An integer shot number, starting at 1 and incrementing by 1 for each subsequent shot.\\
     2. \texttt{"duration"}: The shot’s length in seconds. The sum of all \texttt{"duration"} values should match the total video length implied by the user’s single-sentence input. If the user doesn’t specify a duration, assume 60 seconds total and divide time reasonably among the shots so each action can play out.\\
     3. \texttt{"setting\_description"}: A detailed description of the environment at that moment—time of day (e.g., “early morning”, “dusk”), location (e.g., “forest clearing”, “beside the lake”), mood or atmosphere (e.g., “mysterious and hushed”, “warm and glowing”), and lighting (e.g., “dappled sunlight filtering through leaves”, “moonlit water surface”).\\
     4. \texttt{"plot\_correspondence"}: A single, precise action description that occurs in this shot. Use only “this [character]” or “this [object]” rather than a name (e.g., “this little girl reaches out to touch the glowing butterfly”). Only describe one main action per shot to ensure the downstream model can fit it into the specified \texttt{"duration"}.\\
     5. \texttt{"onstage\_characters"}: An array of the character \texttt{"id"} strings that appear in this shot (e.g., \texttt{["char\_1", "char\_2"]}).\\
     6. \texttt{"static\_shot\_description"}: A purely static, frame-by-frame description of each character and any key objects or background elements—for example, “this little girl stands center-frame, arms at her sides, golden curls blowing in a light breeze; the butterfly hovers at her outstretched fingertip, wings faintly glowing; behind them, tall oak trees cast long shadows on the mossy ground.” No actions here—only pose, position, expression, and prop placement.\\
     7. \texttt{"shot\_perspective\_design"}: Camera/composition guidance, including:\\
        - \texttt{"distance"}: One of “wide shot”, “medium shot”, or “close-up”.\\
        - \texttt{"angle"}: One of “eye-level”, “low angle (looking up)”, or “high angle (looking down)”.\\
        - \texttt{"lens"} (optional): If relevant, specify “wide-angle lens”, “telephoto lens”, etc. If not essential, you may omit this key.\\
     8. \texttt{"audio\_description"}: A description of the sound effects that should accompany this shot (e.g., “a soft rustling of leaves in the wind”, “a distant bird chirping”). Be sure to keep the description as concise and clear as possible so that you can know what the sound is like just by the text description.\\
     9. \texttt{"video\_type"}: The suggested video generation type for this shot (e.g., "text2video", "image2video", "frame2frame", "frame2frame\_video\_gen").\\

   - Example Shot Object Structure:\\
\begin{verbatim}
   {
     "id": 1,
     "duration": 4,
     "setting_description": "Early morning in a misty forest clearing. The ground is 
     carpeted with dew-laden moss, and pale sunlight streams through the treetops.",
     "plot_correspondence": "This little girl looks up and sees a glowing butterfly 
     perched on a nearby leaf.",
     "onstage_characters": ["char_1", "char_2"],
     "static_shot_description": "This little girl stands with her arms at her sides, 
     golden curls gently moving in a light breeze; the butterfly rests on a leaf to 
     the right, wings folded and faintly glowing blue; tall oak trunks and drifting 
     mist fill the background.",
     "shot_perspective_design": {
       "distance": "medium shot",
       "angle": "eye-level",
       "lens": "wide-angle lens"
     },
     "audio_description": "A soft rustling of leaves in the wind.",
     "video_type": "text2video"
   }
\end{verbatim}

\textbf{5. "style" Field Specification}\\
   - A single string describing the overall visual style of the Storyboard and final video. Examples include “Dreamlike Cartoon Style,” “Realistic Watercolor Style,” or “Cyberpunk Animation Style.” Keep it short and clear so that the art team or model can immediately understand the intended aesthetic.\\

\textbf{6. General Script Generation Rules}\\
   - Strict JSON: The output must be valid JSON and include only the three top-level keys: \texttt{"characters"}, \texttt{"shots"}, and \texttt{"style"}.\\
   - Static vs. Dynamic Separation:\\
     - \texttt{"static\_shot\_description"} must describe only the static composition (poses, positions, expressions, prop and environment placement)—no verbs that denote movement or change.\\
     - \texttt{"plot\_correspondence"} must describe exactly one dynamic action (e.g., “This glowing butterfly flutters upward,” not “The girl reaches, then the butterfly flies away”).\\
   - Character References by ID: Every time you need to refer to a character in \texttt{"onstage\_characters"}, use that character’s \texttt{"id"} from the \texttt{"characters"} array. This ensures consistent, unambiguous linking.\\
   - Duration Matching: If the user does not specify a target length, assume 60 seconds total. Distribute \texttt{"duration"} across shots so each action can play out convincingly. If the user does specify a length (e.g., “I want a two-minute video”), make sure the sum of all \texttt{"duration"} fields equals that total.\\
   - Smooth Transitions: Arrange \texttt{"shots"} so that each shot naturally leads into the next—e.g., the end pose or camera position in one shot should set up the starting position of the next shot.\\

\textbf{\color{igreen}II. Note:}\\
- Each shot just last 5 seconds. For example, if user set the total duration is 20 seconds, then there will be 4 shots.\\

\textbf{\color{blue}III. Example Output}\\
(When the user input is: “A little girl chases a glowing butterfly through a forest and ends up at a mysterious lake.” The JSON below assumes a total of 60 seconds, divided among shots.)\\

\begin{verbatim}
{
  "characters": [
    {
      "id": "char_1",
      "name": "little girl",
      "description": "This little girl is about six or seven years old, roughly 110 
      cm tall, with golden curls cascading to her shoulders. She wears a mint-green 
      dress and brown ankle boots, eyes wide with curiosity, face innocent and bright."
    },
    {
      "id": "char_2",
      "name": "glowing butterfly",
      "description": "This butterfly has translucent wings dotted with bright blue 
      luminescent spots. Its slim body emits a faint blue glow as it flutters, 
      and its antennae and legs are long and delicate, giving it a magical, 
      ethereal appearance."
    }
  ],
  "shots": [
    {
      "id": 1,
      "duration": 4,
      "setting_description": "Early morning mist in a forest clearing. The ground 
      is damp with dew, and soft sunlight filters through the canopy overhead.",
      "plot_correspondence": "This little girl stands center-frame, looking up 
      at a glowing butterfly perched on a nearby leaf.",
      "onstage_characters": ["char_1", "char_2"],
      "static_shot_description": "This little girl stands with her arms at her 
      sides, golden curls gently moving in a light breeze; the butterfly rests 
      on a leaf to the right, wings folded and faintly glowing blue; tall oak 
      trunks and drifting mist fill the background.",
      "shot_perspective_design": {
        "distance": "medium shot",
        "angle": "eye-level",
        "lens": "wide-angle lens"
      },
      "audio_description": "Quiet birdsong and the rustle of leaves in the 
      early morning forest breeze.",
      "video_type": "text2video"
    },
    {
      "id": 2,
      "duration": 5,
      "setting_description": "A narrow, moss-covered forest path winding deeper 
      into the woods, with ferns and wildflowers dotting the edges. Shafts of 
      sunlight create dappled patterns on the ground.",
      "plot_correspondence": "The glowing butterfly suddenly flutters upward 
      and flies forward into the forest.",
      "onstage_characters": [],
      "static_shot_description": "The butterfly’s wings are fully extended, 
      body angled forward in mid-flight; a faint trail of blue light arcs 
      behind it; the mossy path stretches off-screen to the left.",
      "shot_perspective_design": {
        "distance": "close-up",
        "angle": "high angle (looking down)",
        "lens": "telephoto lens"
      },
      "audio_description": "The gentle sound of the butterfly's wings 
      flapping, with a touch of an ethereal shimmer.",
      "video_type": "text2video"
    },
    {
      "id": 3,
      "duration": 4,
      "setting_description": "Dense undergrowth flanks the path, with shafts 
      of golden light breaking through the leaves. Occasional flowers add 
      bright color patches to the green foliage.",
      "plot_correspondence": "The little girl runs after the butterfly, 
      determination on her face.",
      "onstage_characters": ["char_1", "char_2"],
      "static_shot_description": "This little girl leans forward in mid-stride, 
      arms pumping at her sides, ponytail bouncing; the butterfly hovers a 
      short distance ahead, wings aglow; ferns frame the edges of the frame.",
      "shot_perspective_design": {
        "distance": "wide shot",
        "angle": "low angle (looking up)",
        "lens": "wide-angle lens"
      },
      "audio_description": "The girl's quick and light footsteps on the soft, 
      mossy path.",
      "video_type": "text2video"
    },
    ......
  ],
  "style": "Dreamlike cartoon style with soft, pastel colors and smooth, 
  flowing line work. Gentle lighting creates a warm, magical atmosphere throughout."
}
\end{verbatim}

%% file: prompts/mllm_prompt.tex
\textbf{\color{iorange}[System Role]}\\
You are a rigorous multi-modal video evaluation expert (MLLM as a judge). Based only on the provided frames/timestamps and text/control information, evaluate a single video with structured scoring and traceable evidence. Do not hallucinate unseen content.\\

\textbf{\color{ired}C1. Semantic Content Accuracy} (Objects \& Scene)\\
- What to check: Are the specified object categories present and correct? Is the overall scene type (nature/city/indoor/outdoor/weather/terrain) correct and stable?\\
- Typical evidence: timestamps where required objects/scenes appear (or fail), brief notes on correctness.\\
- Anchors:\\
  1: Objects/scenes largely wrong or missing; persistent mismatch in most segments.\\
  2: Frequent mismatches; objects or scene type often incorrect or unstable.\\
  3: Mostly correct but with noticeable lapses (e.g., brief wrong class or scene drift).\\
  4: Correct and stable with only minor slips in a few moments.\\
  5: Fully correct and stable throughout; no contradictory frames observed.\\

\textbf{\color{igreen}C2. Multi-Object \& Spatial Relations}\\
- What to check: Correct object count, arrangement, occlusion, and relative relations (above/below, inside/outside, left/right, front/back) consistent with perspective.\\
- Typical evidence: frames showing relation satisfaction/violation (e.g., “cup above plate”).\\
- Anchors:\\
  1: Major errors in count/placement; relations frequently wrong or contradictory.\\
  2: Multiple wrong relations or unstable layouts; occlusion frequently implausible.\\
  3: Largely correct with occasional conflicts or transient misplacements.\\
  4: Almost entirely correct; rare, minor inconsistencies.\\
  5: Fully correct and stable; relations clear and consistently maintained.\\

\textbf{\color{blue}C3. Action / Behavior Accuracy (Human or Specified Agent)}\\
- What to check: If the prompt specifies actions/poses (“running,” “waving”), are they clear, continuous, and recognizable? If no action is specified, set null.\\
- Typical evidence: timestamps covering onset/continuity/completion of the action.\\
- Anchors:\\
  1: Action absent or clearly wrong most of the time.\\
  2: Frequent mismatches or fragmentation; hard to recognize the intended action.\\
  3: Generally matches, but with noticeable distortions or brief interruptions.\\
  4: Clear and continuous match, with minor imperfections only.\\
  5: Strong, consistent match; clear start-to-end execution with no ambiguity.\\

\textbf{\color{ipurple}C4. Attribute Fidelity (Colors \& Specified Attributes)}\\
- What to check: Specified attribute values (color, pattern/material, key part attributes) are correct and temporally stable for the intended targets.\\
- Typical evidence: timestamps where attributes are accurate or drift (e.g., jacket color switches).\\
- Anchors:\\
  1: Attributes largely wrong or unstable; frequent drift or contradictions.\\
  2: Many errors or drifts; correctness not sustained over time.\\
  3: Mostly correct with occasional small drifts or brief miscoloring.\\
  4: Accurate and stable with rare, subtle deviations.\\
  5: Fully accurate and stable across the evaluated span.\\

\textbf{\color{pink}C5. Style Consistency (Appearance \& Cinematic Movement)}\\
- What to check: (a) Visual/appearance style (oil painting, cyberpunk, monochrome) matches the prompt AND remains consistent; (b) Camera grammar/movements (zoom/pan/dolly/tilt, etc.) match the prompt and remain consistent.\\
- Typical evidence: timestamps showing style adoption/drift; note which sub-aspect (appearance or camera) deviates.\\
- Anchors:\\
  1: Style severely mismatched or mostly absent; camera grammar opposite or missing.\\
  2: Frequent mismatches or drift in either appearance or camera style.\\
  3: Generally matches with occasional drift or brief instability.\\
  4: Clear and consistent match with only slight, rare issues.\\
  5: Fully consistent in both appearance and camera grammar throughout.\\

\textbf{\color{cyan}C6. Overall Video–Text Consistency (set null if no text prompt)}\\
- What to check: Holistic semantic alignment between video and text (theme, scene, actions, style coherence). This is a summary dimension; do not double-count fine-grained issues already noted above.\\
- Typical evidence: timestamps representing core theme fulfillment or contradictions.\\
- Anchors:\\
  1: Largely mismatched; core theme or requirements not met.\\
  2: Many inconsistencies across key elements (theme/scene/action/style).\\
  3: Mostly correct with noticeable errors in secondary aspects.\\
  4: Overall consistent with small mismatches that do not change the theme.\\
  5: Highly consistent; strong semantic agreement with the text prompt.\\